\documentclass[final]{cvpr}

\usepackage{times}
\usepackage{epsfig}
\usepackage{graphicx}
\usepackage{amsmath}
\usepackage{amssymb}

\usepackage[utf8]{inputenc}
\usepackage{multirow}
\usepackage[caption=false]{subfig}
\usepackage{verbatim}
\usepackage{wrapfig}
\usepackage{booktabs}
\usepackage{array}
\usepackage{gensymb}
\usepackage{diagbox}
\usepackage[T1]{fontenc}    
\usepackage{url}            
\usepackage{booktabs}       
\usepackage{amsfonts}       
\usepackage{nicefrac}       
\usepackage{microtype}      
\usepackage{setspace}
\usepackage{multirow}
\usepackage{amssymb}
\usepackage{diagbox}
\usepackage{subfig}
\usepackage{wrapfig}
\usepackage{diagbox} 
\usepackage{wrapfig,lipsum,booktabs}
\usepackage{caption}

\usepackage{color}
\usepackage{xcolor}
\definecolor{citecolor}{HTML}{0071bc}
\usepackage[pagebackref=false,breaklinks=true,letterpaper=true,citecolor=citecolor,colorlinks,bookmarks=false]{hyperref}

\definecolor{tblue}{RGB}{80,80,245}
\definecolor{tred}{RGB}{250,100,100}

\newcolumntype{x}[1]{>{\centering\arraybackslash}p{#1pt}}

\newcommand{\app}{\raise.17ex\hbox{$\scriptstyle\sim$}}

\newlength\savewidth\newcommand\shline{\noalign{\global\savewidth\arrayrulewidth
  \global\arrayrulewidth 1pt}\hline\noalign{\global\arrayrulewidth\savewidth}}
\newcommand{\tablestyle}[2]{\setlength{\tabcolsep}{#1}\renewcommand{\arraystretch}{#2}\centering\footnotesize}

\newcommand{\customfootnotetext}[2]{{
  \renewcommand{\thefootnote}{#1}
  \footnotetext[0]{#2}}}

\usepackage{amssymb}
\usepackage{pifont}
\newcommand{\cmark}{\ding{51}}%
\newcommand{\xmark}{\ding{55}}%

\def\cvprPaperID{6336} 
\def\confYear{CVPR 2021}

\begin{document}

\title{Semi-Supervised 3D Hand-Object Poses Estimation \\with Interactions in Time}

\author{
Shaowei Liu\textsuperscript{*}$^{1}$ \quad 
Hanwen Jiang\textsuperscript{*}$^{1}$ \quad
Jiarui Xu$^{1}$ \quad
Sifei Liu$^{2}$ \quad
Xiaolong Wang$^{1}$ \\
$^{1}$ UC San Diego \quad
$^{2}$ NVIDIA
}

\twocolumn[{%
\vspace{-1em}
\maketitle
\vspace{-1em}
\begin{center}
    \includegraphics[width=\linewidth]{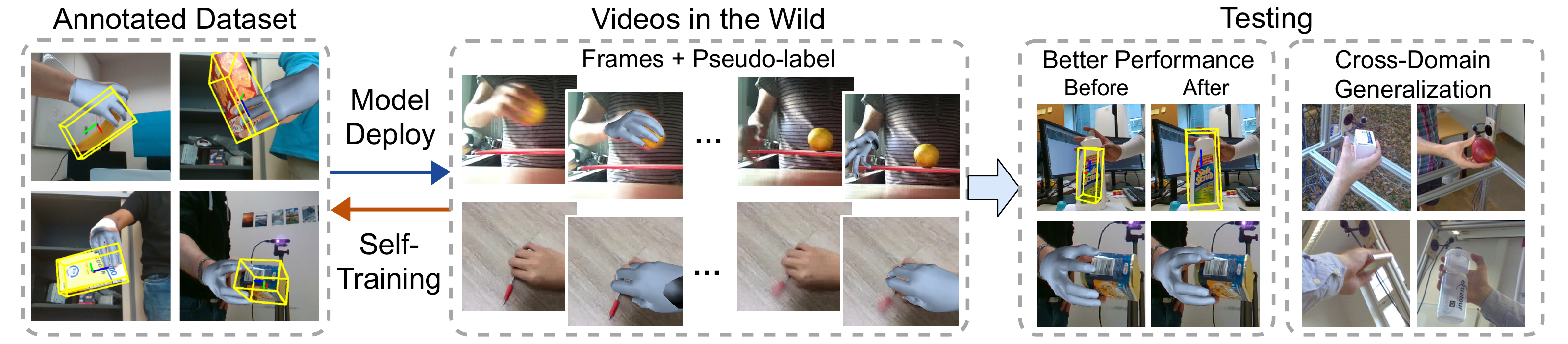}
    \vspace{-0.2in}
    \captionof{figure}{\small{
    The proposed semi-supervised learning framework and results. We train an initial model on an annotated dataset, and deploy it on a large-scale video dataset to collect pseudo-labels. We leverage spatial-temporal consistency to select pseudo-labels for self-training. After the semi-supervised learning, both the performance of hand-object pose estimation and hand pose generalization are improved.}}
    \label{fig:main}
\end{center}
}]
\customfootnotetext{*}{Equal contribution.}

\pagestyle{empty}
\thispagestyle{empty}

\begin{abstract}
   Estimating 3D hand and object pose from a single image is an extremely challenging problem: hands and objects are often self-occluded during interactions, and the 3D annotations are scarce as even humans cannot directly label the ground-truths from a single image perfectly. To tackle these challenges, we propose a unified framework for estimating the 3D hand and object poses with semi-supervised learning. We build a joint learning framework where we perform explicit contextual reasoning between hand and object representations by a Transformer. Going beyond limited 3D annotations in a single image, we leverage the spatial-temporal consistency in large-scale hand-object videos as a constraint for generating pseudo labels in semi-supervised learning. Our method not only improves hand pose estimation in challenging real-world dataset, but also substantially improve the object pose which has fewer ground-truths per instance. By training with large-scale diverse videos, our model also generalizes better across multiple out-of-domain datasets. Project page and code: \href{https://stevenlsw.github.io/Semi-Hand-Object}{https://stevenlsw.github.io/Semi-Hand-Object} .
\end{abstract}

\section{Introduction}

Hands are humans' primary means of interacting with the physical world. Capturing the 3D pose of the hands and the objects interacted by hands is a crucial step in understanding human actions. It is also the central part for a variety of applications including augmented reality~\cite{piumsomboon2013user,hurst2013gesture}, third-person imitation learning~\cite{handa2020dexpilot,garcia2020physics}, and human-machine interaction~\cite{sridhar2015investigating}. While 3D pose estimation on hands and objects have been studied for a long time in computer vision captured with depth cameras~\cite{yuan2018depth, moon2018v2v, 9028217, xu2017lie, li2020category} or RGB-D sensors~\cite{yao2012real, rogez20143d, mueller2017real, schwarz2015rgb, GarciaHernando2018FirstPersonHA} in controlled environments, recent research has also achieved encouraging results on pose estimation from a single monocular RGB image~\cite{zimmermann2017learning,GANeratedHands_CVPR2018,xiang2017posecnn}.

Despite the efforts, current approaches still highly rely on human annotations for 3D poses, which are extremely difficult to obtain: Researchers have been collecting data with motion capture~\cite{romero2017embodied, glauser2019interactive, zhou2020monocular}, or aligning mesh models to the real images~\cite{hasson2019learning, dkulon2020cvpr, Hampali2019HOnnotateAM, boukhayma20193d}. Given insufficient annotations for supervised learning, it limits the trained model from generalizing to novel scenes and out-of-domain environments. To enable better estimation performance and generalization ability, we look into video data of hands and objects in the wild, without using the 3D annotations.

Specifically, we propose to exploit hand-object interactions over time. The poses of the hands and objects are usually highly correlated: The 3D pose of the hand when it is grasping the object often indicates the orientation of the object; the object pose also provides constraints on how the hand can approach and interact with the object. 

When observing from the videos, the 3D poses for both hands and objects should change smoothly and continuously. This continuity provides a cue for selecting coherent and accurate 3D hand and object pose estimation results when human annotations are not available. 

In this paper, we introduce a semi-supervised learning approach for 3D hand and object pose estimation with videos.
We first train a joint model for both 3D hand pose and 6-Dof object pose estimation with supervised learning using fully annotated data. Then we deploy the model for hand pose estimation in large-scale videos without 3D annotations. We collect the estimation results as novel pseudo-labels for self-training. 
Specifically, to utilize the interaction information between hand and object, we design a unified framework that extracts the representation from the whole input image, and uses RoIAlign~\cite{he2017mask} to further obtain the object and hand region representations.
Building on these representations, we apply two different branches of sub-networks to estimate the 3D poses for hand and object, respectively. We use the Transformer~\cite{vaswani2017attention} to bridge the two branches for encoding the mutual context between hand and object. The attention mechanism in Transformer could well model the their relations at every pixel location and allow us to take hand-object interaction as an advantage instead of a disadvantage.

To perform semi-supervised learning with hand-object videos, we deploy our unified model on each frame for pseudo-label generation, as illustrated in Figure~\ref{fig:main}. Given the 3D hand pose results from our model, we design spatial-temporal consistency constraints to filter unstable and inaccurate estimations. Intuitively, we only keep the results as pseudo-labels if they change continuously over time, which indicates the robustness of the estimation. We then perform self-training with the newly collected data and labels.

We experiment by training the initial model on the HO-3D dataset~\cite{Hampali2019HOnnotateAM}, and perform semi-supervised learning with the Something-Something video dataset~\cite{Goyal2017TheS}. By learning from the pseudo-labels from large-scale videos using our approach, we achieve a large gain over state-of-the-art approaches in the HO-3D benchmark. We also show significant improvements in 3D hand pose estimation which generalizes to the out-of-domain datasets including FPHA~\cite{GarciaHernando2018FirstPersonHA} and FreiHand~\cite{zimmermann2019freihand} datasets. More surprisingly, even though we only use pseudo-labels for hands, our joint self-training improves the object pose estimation by a large margin (more than $10\%$ in some objects). 

Our contributions include: (i) An unified framework for joint 3D hand and object pose estimation; (ii) A semi-supervised learning pipeline which exploits large-scale unlabeled hand-object interaction videos; (iii) Substantial performance improvement on hand and object pose estimation, and generalization to out-of-domain data.

\begin{figure*}[t]
\centering
\includegraphics[width=17.5cm]{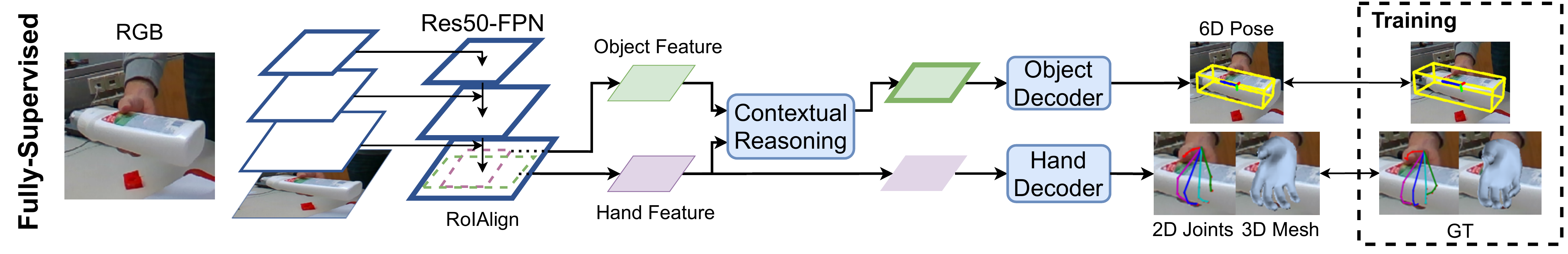}
\vspace{-0.25in}
\caption{\small{Overview of the framework for hand-object 3D poses joint estimation. The model consists of the shared Res-50-FPN encoder, the transformer-based contextual reasoning module for modeling hand-object interaction, and two decoders for estimating hand-object poses. The hand decoder outputs 2D joints and 3D mesh parameterized by the MANO model. The object decoder predicts 2D location of pre-defined 3D control points on the object and recovers 6-Dof pose by the PnP algorithm. 
}}
\vspace{-0.1in}
\label{fig: pipeline}
\end{figure*}

\section{Related Work}

\textbf{Hand pose estimation.} 
Research in RGB-based hand pose estimation can be generally categorized into two paradigms, model-free approaches~\cite{cai2018weakly, zimmermann2017learning, iqbal2018hand, spurr2018cvpr, GANeratedHands_CVPR2018, ge20193d} and model-based approaches~\cite{baek2019pushing, boukhayma20193d, dkulon2020cvpr, zhang2019end, hasson2019learning}. 
Model-free approaches estimate pose by learning joint coordinates~\cite{zimmermann2017learning, spurr2018cvpr} or joint heatmaps~\cite{iqbal2018hand, GANeratedHands_CVPR2018, cai2018weakly}. For example, Zimmermann \textit{et al.}~\cite{zimmermann2017learning} proposed to detect 2D hand joints and lift them into 3D with the articulation prior. Differently, model-based approaches utilize the differentiable MANO model~\cite{romero2017embodied} to capture hand pose and shape. For example, Boukhayma~\textit{et al.}~\cite{boukhayma20193d} collected synthetic data for pre-training to increase the hand pose accuracy. In our work, instead of relying on either synthetic data or 3D ground-truths, we leverage the spatial-temporal information in large-scale real-world videos to achieve better hand pose estimation performance and generalization ability in a semi-supervised manner. 

\textbf{Object pose estimation.} There are also two main paradigms to perform object 6-Dof pose estimation, with one directly regressing the pose as network outputs~\cite{Kehl2017SSD6DMR, Xiang2018PoseCNNAC} and another regressing the projected 3D object control points location in the image and recovering the pose with 2D-to-3D correspondence~\cite{Rad2017BB8AS, Tekin2018RealTimeSS, Peng2019PVNetPV, Hu2019SegmentationDriven6O}. Due to the non-linearity of the rotation space, direct regression of the 6-D pose suffers from the generalization problem~\cite{Hu2019SegmentationDriven6O, Xiang2020RevisitingTC}. For the 2D-to-3D genre, as an example,  Hu \textit{et al.}~\cite{Hu2019SegmentationDriven6O} performed inference by generating pixel-wise structural outputs, containing multiple proposals for computing the pose, which demonstrates strong robustness and efficiency. Our method is inspired by this approach, but further extends to consider the hand-object interaction for object pose estimation. The Transformer architecture in our joint framework performs contextual reasoning to enhance the object features and lead to better object pose estimation under occlusion.

\textbf{Hand-object interaction.}
Simultaneously estimating hand and object poses~\cite{tekin2019h+, choi2017robust, oberweger2019generalized, oikonomidis2011full, hasson2020leveraging, Doosti2020HOPENetAG, Chen2020JointH3} during interaction is a challenging task due to self-occlusion. To tackle this problem, Hasson~\textit{et al.}~\cite{hasson2019learning} leveraged physical constraints for regressing hand and object mesh at the same time using two separate networks. Differently, we observe that sharing the feature backbone in learning between two pose estimation tasks can implicitly encode the context information. And this contextual information becomes very useful when applied to the semi-supervised learning setting. Similar to our approach, Chen~\textit{et al.}~\cite{Chen2020JointH3} proposed to fuse hand-object representations to get interaction-aware features for joint pose estimation using LSTM~\cite{Greff2017LSTMAS}. However, this feature fusion method cannot model the spatial dependency between hand and object. Instead, our method utilizes a Transformer to perform contextual reasoning between hand-object representations to get interaction-aware feature maps explicitly, while maintaining spatial information. The module could benefit both hand and object pose estimation. 

\textbf{Semi-supervised Learning.} Semi-supervised learning plays a key role in improving model performance when the labeled data is limited~\cite{radosavovic2018data, lopez2015unifying, xie2019self,scudder1965probability, yarowsky1995unsupervised, riloff2003learning, rosenberg2005semi, chen2013neil}. Given a trained model on human-annotated datasets, we can apply it on unlabeled data to collect pseudo-labels for further training~\cite{lee2013pseudo, iscen2019label, arazo2019pseudo, shi2018transductive}. For example, Hinton \textit{et al.}~\cite{Hinton2015DistillingTK} have proposed to perform model ensembles in testing to improve the estimation performance. Instead of using multiple models, Radosavovic \textit{et al.}~\cite{radosavovic2018data} proposed to deploy the trained model with test-time augmentations to increase the confidence of the pseudo-labels. While related to our method, most of the previous approaches have not considered the spatial and temporal constraints in videos for selecting the pseudo-labels, which is one of our innovations.

\textbf{Interaction Reasoning.} Interaction reasoning~\cite{Wang2018NonlocalNN, santoro2017simple, watters2017visual, vaswani2017attention, wang2018videos, wu2019long, sadeghian2017tracking, battaglia2016interaction, hu2018relation} aims to model the interactions among objects. For example, Santoro~\textit{et al.}~\cite{santoro2017simple} inferred relations across all pairs of objects to solve the visual question answering task. Wang~\textit{et al.}~\cite{Wang2018NonlocalNN} captured long-range dependencies via the non-local module in spacetime for video classification. Recent studies show the superiority of the Transformer~\cite{vaswani2017attention} architecture in learning visual relations and using it to solve different computer vision tasks~\cite{parmar2018image, ramachandran2019stand, sun2019videobert, lee2019set, dosovitskiy2020image, carion2020end, chi2020relationnet++}, especially for pose estimation~\cite{ yang2020transpose, mao2021tfpose, stoffl2021end, huang2020hand, lin2021end-to-end, hampali2021handsformer}. The goal of our work is to exploit the visual correlation between hand and object to improve pose estimation performance under occlusion. We adopt the Transformer architecture in the contextual reasoning module, where we do cross-attention to exploit the relevant pair of cells between hand-object instead of the whole image.

\section{Overview}

Our method on 3D hand and object pose estimation contains two main components: (i) a joint learning framework with contextual reasoning between the hand and the object; (ii) a semi-supervised pipeline which explores extra labels in videos for training. 

First, we present the hand-object joint learning framework in Section~\ref{sec:model}. 
The model contains a shared encoder and two separate decoders for hand and object, as well as a transformer-based contextual reasoning module~\ref{sec:cr} to better exploit their relations. The model is trained under fully-supervised learning. 

Then, we propose the semi-supervised learning pipeline in Section~\ref{sec:semisup},   Constrained by the spatial-temporal consistency, we generate high-quality 
pseudo-labels of hand on a large-scale video dataset~\cite{Goyal2017TheS} and re-train our model on the union of fully annotated dataset~\cite{Hampali2019HOnnotateAM} and those confident pseudo-labels. Because of the diversity in the hand pseudo-labels, the model could both increase the accuracy of hand pose estimation and generalization. With better hand features as context via the contextual reasoning module, the object pose performance of the model could also be improved.

\section{Hand-Object Joint Learning Framework}
\label{sec:model}
Our hand-object joint learning framework is presented in Figure~\ref{fig: pipeline}. 
We use FPN~\cite{Lin2017FeaturePN} with ResNet-50~\cite{He2016DeepRL} as the backbone and
extract hand and object features $\mathcal{F}^h$ and $\mathcal{F}^o$ into $\mathbb{R}^{H\times W\times C}$ with the RoIAlign~\cite{he2017mask}, given their corresponding bounding boxes. Then we apply the contextual reasoning between the two features and send the enhanced feature maps with strengthened interactive context information into the hand and the object decoders respectively, which output the 3D hand mesh and 6-Dof object pose. 
The total loss function of the system is the sum from two decoder branches $\mathcal{L} = \mathcal{L}_{hand} + \mathcal{L}_{object}$. The contextual reasoning module, hand and object decoders, and training losses $\mathcal{L}_{hand}, \mathcal{L}_{object}$ will be discussed in the following sections. 

\begin{figure}[t]
\centering
\includegraphics[width=8cm]{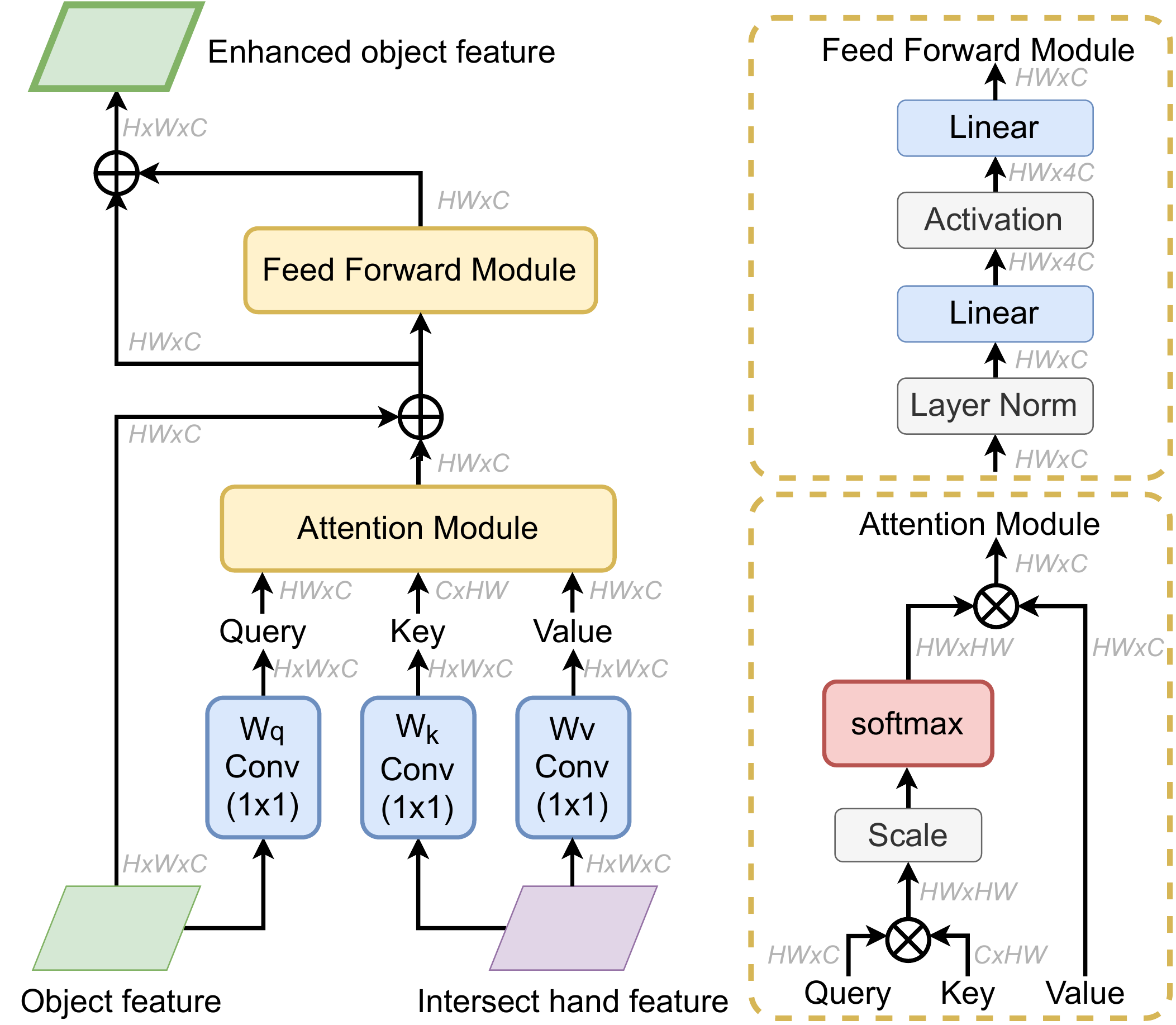}
\caption{\small{The contextual reasoning (CR) module: features extracted from the region of hand intersecting with object are taken as context (key and value) to enhance the object features (query). 
The module adopts the Transformer architecture with an attention module and a feed-forward module. ``$\otimes$'' denotes matrix multiplication, and ``$\oplus$'' denotes element-wise sum. 
}}
\vspace{-0.1in}
\label{fig: contextual}
\end{figure}

\subsection{Contextual Reasoning}
\label{sec:cr}

We adopt the Transformer architecture to exploit the synergy between hand and object features via the contextual reasoning (CR) module as shown in Figure~\ref{fig: contextual}, where the query positions in object features could be enhanced by fusing information from the interaction region. 

Inside the module, we take object features $\mathcal{F}^o$ as query and hand-object intersecting regions $\mathcal{F}^{inter}$ as key and value to model their pairwise relations on the top of RoIAlign~\cite{he2017mask}. We first apply three separate 1-D convolutions  parameterized by $W_q$, $W_k$, and $W_v$ on the input features to get the query, key, and value embedding $Q$, $K$, and $V$:
\begin{align}
Q = W_q \mathcal{F}^h, K = W_k \mathcal{F}^{inter}, V = W_v \mathcal{F}^{inter}
\end{align}
The above equation is also shown on the left side of Figure~\ref{fig: contextual}. 

Then the embeddings are forwarded to the attention module to perform cross-attention between every spatial location of the query and the key. The output of the attention module $\mathrm{Att}(Q, K, V)$ is the weighted average of the value computed as follows:
\begin{align}
    \mathrm{Att}(Q, K, V) = \mathrm{softmax}(\frac{QK^T}{\sqrt{d_k}})V
\end{align}
where $d_k$ is the feature dimension of the key and $\mathrm{softmax}$ stands for the softmax function. The attention module output is fused to the original input to get enhanced query features $Q'$:
\begin{align}
    Q' = \mathcal{F}^o + \mathrm{Att}(Q, K, V) 
\end{align}

Afterward, the obtained features $Q'$ are sent to the feed-forward module that consists of a two-layer MLP and layer normalization~\cite{ba2016layer}, abbreviated as $\mathrm{LN}$. The detailed architecture of the attention module and feed-forward module are shown on the right side of Figure~\ref{fig: contextual}. The final output of the CR module is the enhanced object features $\mathcal{F}^{o^+}$, obtained by adding up the feed-forward module output with residual connections:
\begin{align}
        \mathcal{F}^{o^+} = Q' + \mathrm{MLP}(\mathrm{LN}(Q'))
\end{align}

Besides, we also ablate on using different features as the query of CR module in Section~\ref{sec: ablation_cr}, where we alternatively choose to use the features of hand or both hand-object as the query.

\subsection{Hand Decoder}
The hand decoder consists of a 2D joints localization network and a mesh regression network. The 2D joints localization network is an hourglass~\cite{newell2016stacked} module which takes the hand features after RoIAlign~\cite{he2017mask} as input and outputs 2D heatmaps for each joint $j \in \mathcal{J}^{2D}$, where $\mathcal{J}^{2D} \in \mathbb{R}^{N^h \times 2}$ and $N^h = 21$ is the number of joints. The heatmaps have the resolution $32 \times 32$. The loss function of the 2D joints localization network $\mathcal{L}_{\mathcal{H}}$ is the distance between ground-truth heatmaps $\mathcal{H}_{j}$ and predictions $\hat{\mathcal{H}}_{j}$ of each joint $j$, as
$ \mathcal{L}_{\mathcal{H}} = \sum_{j \in N^h} \vert\vert \mathcal{H}_{j} -\hat{\mathcal{H}}_{j}  \vert\vert_2^2$.

The mesh regression network combines the hand features with the 2D heatmaps as input and predicts the parameters of the hand mesh parameterized by the MANO model~\cite{romero2017embodied}. The MANO model maps the pose parameters $\theta \in \mathbb{R}^{48}$ and shape parameters $\beta \in \mathbb{R}^{10}$ to hand mesh vertices $\mathcal{V} \in \mathbb{R}^{778 \times 3}$ and 3D joints $\mathcal{J}^{3D} \in \mathbb{R}^{N^h \times 3}$. The inputs of the mesh regression network are forwarded to a ConvNet with four residual blocks ~\cite{He2016DeepRL} and vectorized into a 2048-D feature vector. The output are the predicted MANO parameters $\hat{\theta}$ and $\hat{\beta}$ using three fully-connected layers. We compute the loss on both the MANO model parameters and outputs. Specifically, we compute the $L_2$ distance between the prediction $(\hat{\theta}, \hat{\beta}, \hat{\mathcal{J}^{3D}}, \hat{\mathcal{V}})$ and ground-truth $(\theta, \beta, \mathcal{J}^{3D}, \mathcal{V})$ as the loss $L_{\mathcal{M}}$ in mesh regression network. The total loss of the hand decoder is the sum of heatmap loss $\mathcal{L}_{\mathcal{H}}$ and $L_{\mathcal{M}}$:
\begin{align}
    \mathcal{L}_{hand} = \lambda_{\mathcal{H}} \cdot \mathcal{L}_{\mathcal{H}} + \mathcal{L}_{\mathcal{M}} \ .
\end{align}
where $\lambda_{\mathcal{H}}=0.1$ is used for balancing losses.

\subsection{Object Decoder}
The object decoder consists of two streams, which has 4 shared convolution layers and 2 separate convolution layers for each stream. The first stream predicts the 2D location of pre-defined 3D control points on the object from image grid proposals, and the second stream regresses the corresponding confidence scores of each proposal. After obtaining the 2D positions of control points, the object 6-Dof pose can be computed by the PnP algorithm using the correspondence between 2D control points and original 3D control points on the object mesh. In this work, we utilize $N^o = 21$ control points, including $8$ corners, $12$ edge midpoints and $1$ center-point of the object mesh 3D bounding box. 

In the first stream, we adopt the grid-based method~\cite{Redmon2016YouOL} to better handle self-occlusion, where each grid $g$ in the object feature map gives a prediction for every control point $i \in N^o$. We use $\delta_{g, i}$ to denote the geometric distance between the grid prediction and the target control point. The loss function of the first stream is the loss sum over all the grids $g$ and control points $i$, denote as $\mathcal{L}_{p2d} = \sum_{g} \sum_{i=1}^{N^o} \vert\vert \delta_{g, i} \vert\vert_1 $. 

The second stream regresses a confidence score $c_{g,i}$ for each grid $g$ and control point $i$, where the confidence ground-truth $c_{g,i} = {\rm exp} (- \vert\vert \delta_{g,i} \vert\vert_2)$, which indicates the proximity of the prediction to the ground-truth 2D point locations. During test time, we pick 10 most confident proposals as the input of the PnP algorithm to solve for the object pose. The loss function of the second stream is denoted as $\mathcal{L}_{conf} = \sum_{g} \sum_{i=1}^{N^o} \vert\vert \hat{c}_{g,i} - c_{g,i} \vert\vert_2^2$, where $c_{g,i}$ and $\hat{c}_{g,i}$ are the ground-truth and predictions. 

The total loss of the object decoder is:
\begin{align}
    \mathcal{L}_{object} = \lambda_{p} \cdot \mathcal{L}_{p2d} + \lambda_{c} \cdot \mathcal{L}_{conf} \ .
\end{align}
where $\lambda_{p}=0.5$ and $\lambda_{c}=0.1$ are hyperparameters.

\begin{figure}[t]
\centering
\includegraphics[width=8cm]{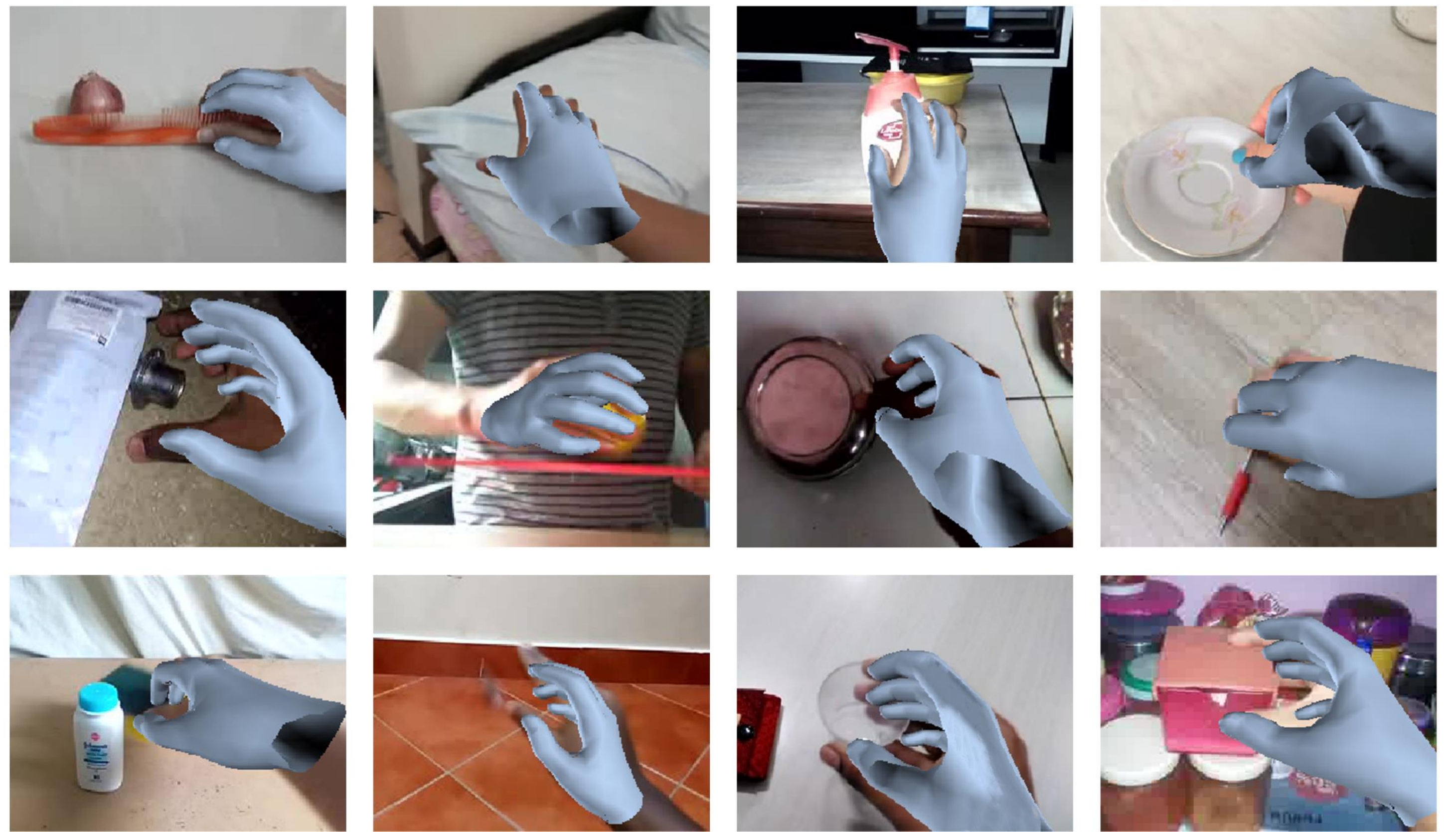}
\vspace{-0.1in}
\caption{\small{Examples of filtered hand mesh pseudo-labels generated from the video dataset~\cite{Goyal2017TheS}.}}
\vspace{-0.1in}
\label{fig:vis_pseudo}
\end{figure}

\section{Semi-Supervised Learning}
\label{sec:semisup}
After we trained the model of hand-object pose estimation on the fully annotated dataset, we deploy it on a large-scale unlabeled video dataset~\cite{Goyal2017TheS} for 3D hand pseudo-label generation. We leverage spatial-temporal consistency to filter out unreliable pseudo hand labels. The obtained pseudo-label examples on the video dataset~\cite{Goyal2017TheS} can be seen in Figure~\ref{fig:vis_pseudo}. Note that we do not generate pseudo-labels for objects because of the need for object 3D models at inference time and the poor generalization of object pose due to the limited instances on the annotated dataset. By enlarging the fully annotated dataset with the selected pseudo-labels, we conduct self-training for both hand and object pose estimation. 

\subsection{Pseudo-Label Generation}
We first deploy our model on video frames from a large-scale video dataset~\cite{Goyal2017TheS} for 3D hand pose estimation. To improve the estimation robustness, we do test-time data augmentation and ensemble the predictions similar to~\cite{radosavovic2018data}. In our experiment, we perform 8 different augmentations of each instance and average the results. The outputs of each frame include 2D joints $\mathcal{J}^{2D}$, 3D joints $\mathcal{J}^{3D}$, 3D hand mesh vertices $\mathcal{V}$, and corresponding MANO parameters $(\theta, \beta)$. 

While ensemble predictions reduce the noise in generated samples, we still need to identify confident ones. To this end, we establish a pipeline for filtering by innovatively utilizing the spatial and temporal consistency constraints in the video dataset, as shown in Figure~\ref{fig: pseudo-label-selecion}.

\vspace{-0.1in}
\subsubsection{Spatial Consistency Constraints} 
\label{sec:st-consist}
\begin{figure}[t]
\centering
\subfloat[Solve the camera projection by utilizing correspondence between estimated 3D hand joints and 2D hand joints. Then project the predicted 3D mesh to 2D.]{
\label{fig: pseudo1}
\includegraphics[width=8cm]{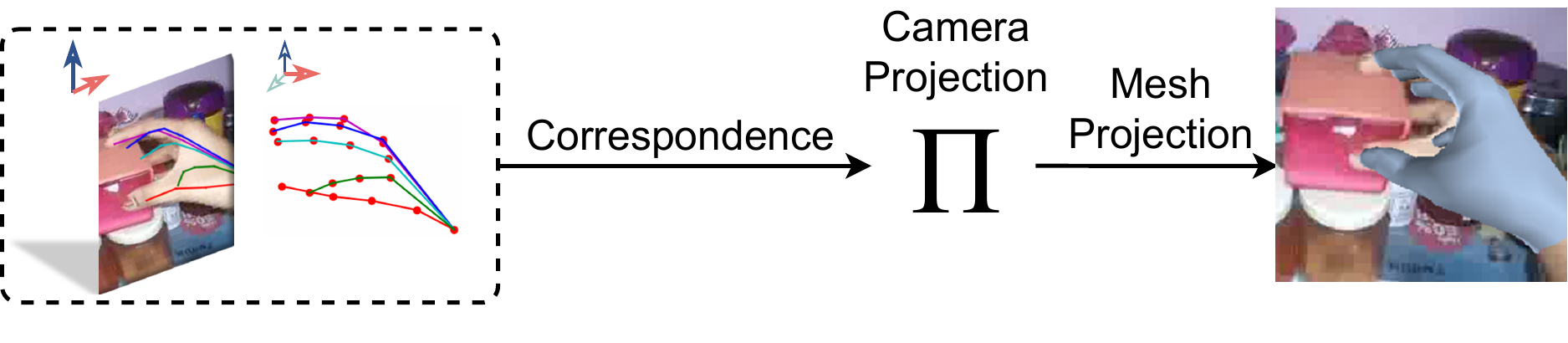}}
\\
\vspace{-0.05in}
\subfloat[Leverage spatial-temporal consistency for pseudo-label selection. First apply spatial constraints in both 2D and 3D, then perform temporal filtering between consecutive frames and within each video sequence.]{
\label{fig: pseudo2}
\includegraphics[width=8cm]{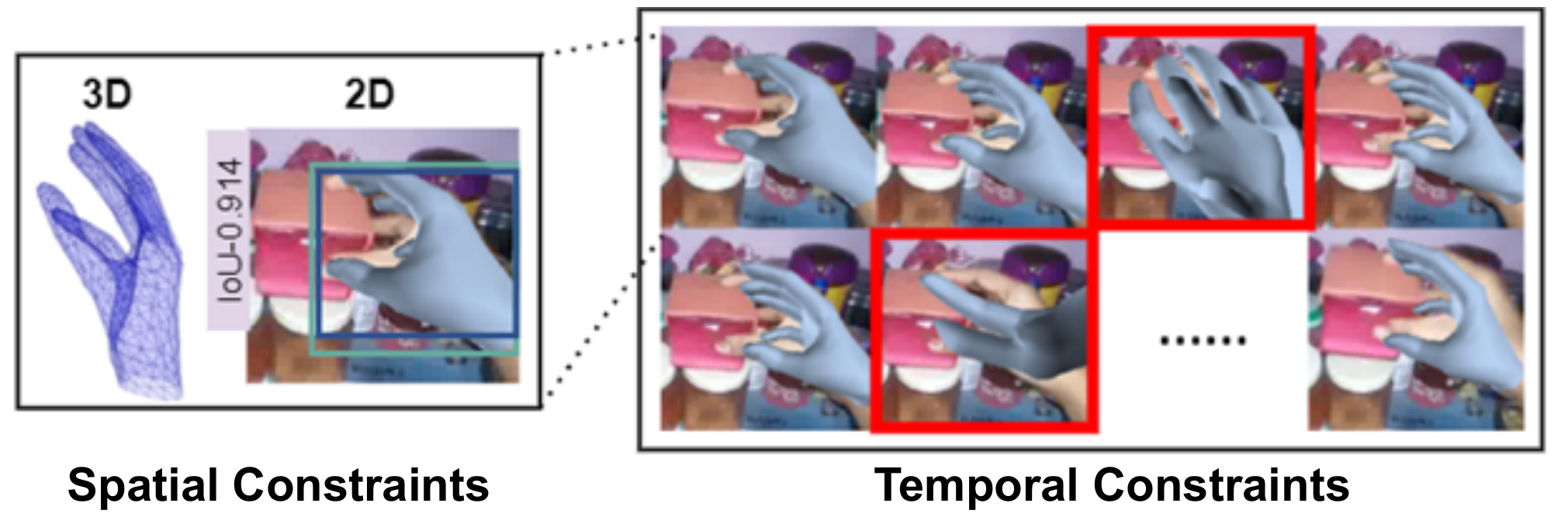}}
\vspace{-0.05in}
\caption{\small{Pipeline of pseudo-label selection for video frames in the wild.}}
\label{fig: pseudo-label-selecion}
\vspace{-0.1in}
\end{figure}

Filtering with spatial consistency requires the corresponding camera pose of each frame. However, it is infeasible to infer the camera pose directly on the video dataset like~\cite{Goyal2017TheS} which has a large variety of viewpoint changes. Our solution to this problem is to leverage the correspondence between the estimated 3D joints $\mathcal{J}^{3D}$ and 2D joints $\mathcal{J}^{2D}$ and solve for the optimal camera parameters $\Pi$ that projects the 3D joints to 2D, as shown in Figure~\ref{fig: pseudo1}. We use the weak-perspective camera model and use the SMPLify~\cite{pavlakos2018learning} for the optimization. The objective is the following: 
\begin{align}
    \Pi^{*} = \arg\min_{\Pi}\vert\vert \Pi\mathcal{J}^{3D}-\mathcal{J}^{2D} \vert\vert_2^2 \ ,
\end{align}
where $\Pi^{*}$ is the optimal camera parameters. 

\textbf{IoU Constraint.} With the camera pose, we can re-project the estimated 3D mesh $\mathcal{V}$ to the image plane and calculate the Intersection-over-Union (IoU) between the provided ground bounding box $B_g$ and the re-projected mesh bounding box $B_d$, as shown in the left side of Figure~\ref{fig: pseudo2}.  Note that although we do not have 3D ground-truths, we leverage the 2D bounding box annotations provided by~\cite{Goyal2017TheS}. A confident prediction should always tend to be consistent between these two boxes and has a large IoU. We set the IoU threshold as $0.6$ for confident prediction.

\begin{figure*}[t]
\centering
\includegraphics[width=17.5cm]{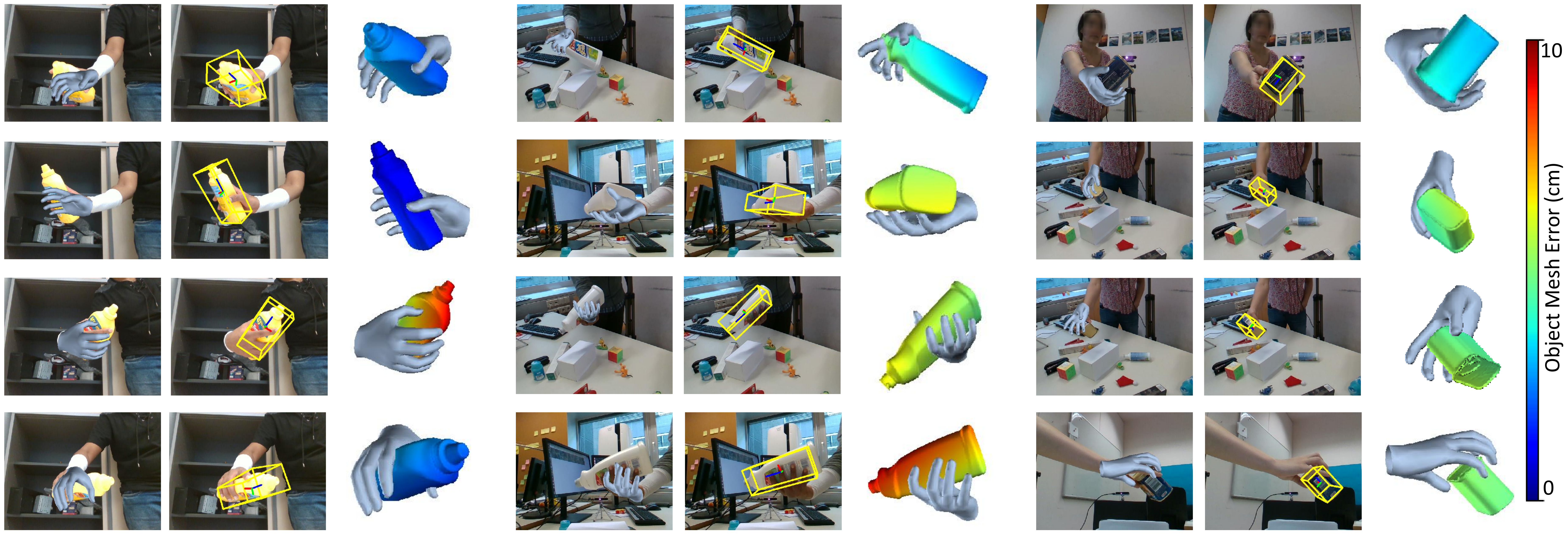}
\vspace{-0.25in}
\caption{\small{Qualitative results of predicted hand-object pose estimation on the HO-3D dataset. For each example, the first two columns show the recovered hand mesh and estimated 6-Dof object pose, the third row shows the estimated hand-object in 3D, the color of the object indicates the 3D object mesh error. Red indicated a larger estimation error. It demonstrates our method can well handle the occlusion in the interaction and recover the accurate 3D hand mesh and object 6-Dof pose.}}
\label{fig:vis_ho3d}
\vspace{-0.1in}
\end{figure*}

\textbf{Pose Re-projection Constraint.} The re-projected 3D hand joints $\Pi \mathcal{J}^{3D}$ and estimated 2D hand joints $\mathcal{J}^{2D}$ should be consistent. We first normalize these two set of joints independent of input sizes and compute the $L_2$ distance between them as $\vert\vert\mathcal{J}^{2D} - \Pi \mathcal{J}^{3D} \vert\vert_2$. When the distance is larger than the threshold $t_p$, then the prediction would be filtered out. We set $t_p=0.65$

\textbf{Biomedical Constraint.} The predicted hand pose should be natural human hands. Thus, we exploit the minimal normalized bone length of $0.1$ and physically plausible joint angle ranges within $(0, 90)$ as two additional constraints to help remove those unnatural predicted hand poses. 

For each instance in the dataset, if it does not violate the three spatial consistency constraints mentioned above, we move forward to the temporal consistency constraints.

\vspace{-0.15in}
\subsubsection{Temporal Consistency Constraints}
\vspace{-0.05in}
\textbf{Smoothness Constraint.} We consider the temporal consistency constraints as the smoothness of both the 2D joint predictions and the 3D mesh predictions between two consecutive frames $t-1$ and $t$. As shown in the right side of Figure~\ref{fig: pseudo2}, since the frames are continuous, the model outputs should be smooth over time. Concretely, the distance of the 2D pose estimation results between these two frames $\vert\vert\mathcal{J}_{t}^{2D} - \mathcal{J}_{t-1}^{2D}\vert\vert_2$ should be less than a threshold $t_j$. Similarly, for the MANO pose parameter $\theta$,  we have $\vert\vert \theta_t - \theta_{t-1} \vert\vert_2 \leq t_{\theta}$ to ensure 3D mesh smoothness,  where $t_j=0.5$ and $t_{\theta}=0.01$ are two constant thresholds.  

\textbf{Shape Constraint.}
In each video sequence, the shape of the hands belongs to the same person should be invariant over time. Given the confident prediction subset $C$ which is the collection of frames satisfying the above constraints in each video sequence, we compute the mean hand shape as $\Bar{\beta} = \frac{1}{\vert C \vert}\sum_{t \in C} \beta_t$ and its standard deviation $\sigma_C = \sqrt{\frac{1}{\vert C \vert}\sum_{t \in C} \vert\vert \beta_t-\Bar{\beta}\vert\vert_2^2)} $. We filter out the frames in $C$ whose shape deviation from $\Bar{\beta}$ is $2$ times larger than $\sigma_C$.

\subsection{Re-training with Pseudo-Labels}
We conduct self-training on the union set of the human-annotated dataset and those pseudo-labels. The diversity of the hand pseudo-labels not only improves the hand prediction, but also provides a richer context for hand-object interaction reasoning via the CR module, leading to better object pose estimation. During the retraining, since we do not have the pseudo-labels of objects, we use a binary mask to ensure only computing the loss of the hand on the pseudo-labeled dataset. The total loss function in the retraining stage is the following:
\begin{align}
    \mathcal{L} = \mathcal{L}_{hand} + \mathcal{B} \cdot \mathcal{L}_{object} \ .
\end{align}
where $\mathcal{B}$ is the mask which equals $1$ on the fully-annotated dataset and $0$ otherwise.

\section{Experiment}
First, we test hand-object pose estimation performance on the HO-3D~\cite{Hampali2019HOnnotateAM} dataset and visualize the prediction in Section~\ref{exp: pose-estimate}. In Section~\ref{exp-ablation}, we conduct abundant ablation studies on the designs of the CR module and explore the effectiveness of semi-supervised learning. In section~\ref{exp-gen}, we test our model's generalization before and after semi-supervised learning on Freihand~\cite{zimmermann2019freihand} and FPHA~\cite{GarciaHernando2018FirstPersonHA} dataset.

\subsection{Implementation Details}
Our model is trained in an end-to-end manner from scratch both in the supervised learning phase and semi-supervised learning phase. The shared encoder in the joint learning framework is initialized with ResNet-50 pre-trained on ImageNet. We use a batch size of 24, initial learning rate $1e^{-4}$,  and Adam optimizer for the training. The training lasts for 60 epochs and the learning rate is scaled by a factor of 0.7 every 10 epochs. We crop the input image to $512 \times 512$ and do data augmentation including scaling $(\pm 20\%)$, rotation $(\pm 180\degree)$, translation $(\pm 10\%)$ and color jittering $(\pm 10\%)$.

\begin{figure}[t]
\centering
\includegraphics[width=8cm]{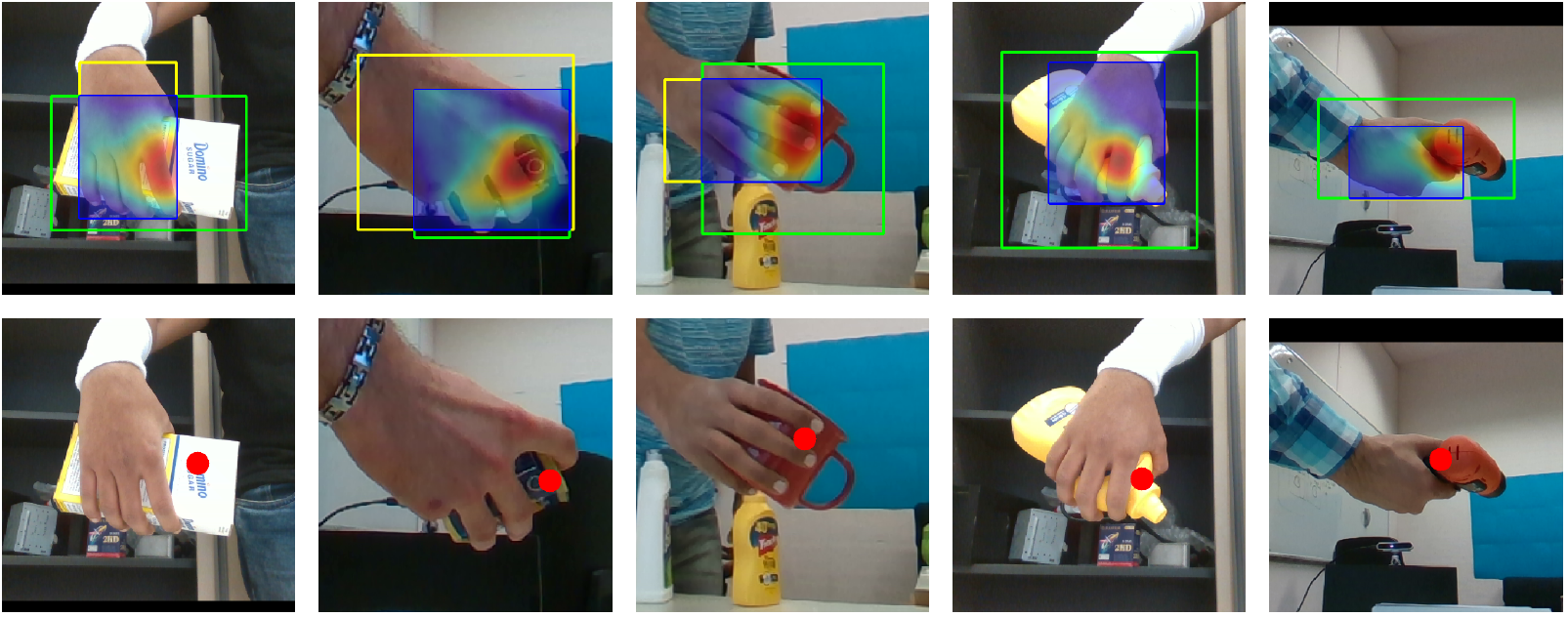}
\vspace{-0.05in}
\caption{\small{Visualization of the CR module. The top row shows the cross-attention weights (red indicates higher response), the bottom row shows corresponding object query positions (red points). Hand, object and intersection boxes are in yellow, green, and blue. The CR module gives high responses to contact regions and tends to use the contact pattern for relational reasoning.}}
\vspace{-0.1in}
\label{fig: vis_att}
\end{figure}

\subsection{Datasets}
\textbf{HO-3D Dataset}~\cite{Hampali2019HOnnotateAM} is used to train our model in supervised learning. The dataset contains more than 65 sequences captured with 10 different subjects and 10 objects with both 3D pose annotations of hand and object. It has $66,034$ and $11,524$ hand-object interaction images from a third-person view for training and testing. The test results are evaluated by the official online submission system.

\textbf{Something-Something Dataset}~\cite{Goyal2017TheS} is a large-scale hand-object interaction video dataset where we conduct semi-supervised learning with only hand and object bounding boxes provided. It covers a variety of hand instances and most daily objects. We finally select $84063$ frames with pseudo hand labels. 

\textbf{FPHA, Freihand Dataset}~\cite{GarciaHernando2018FirstPersonHA, zimmermann2019freihand} are used for validating cross-domain generalization. FPHA is an egocentric video dataset with hand-object interactions. The dataset is captured by using magnetic sensors strapped on hands. The first-person viewpoint and sensors on the hand introduce great challenges for the model's generalization. We use the same subset as previous work~\cite{tekin2019h+, hasson2019learning, hasson2020leveraging} for testing. The test set of the Freihand dataset contains 3960 samples with hands in both indoor and outdoor environments. Hands in half of the samples are interacting with the objects, while in the other half objects are absent. It is also a challenging benchmark for models trained only on hand-object interactions. The evaluation is also performed at the online server.

\subsection{Evaluation Metrics}
For hand pose estimation, We report the standard metric returned from the submission system, i.e. mean joint error and mesh error in mm after Procrustes alignment and F-scores. To further evaluate cross-dataset generalization, we report the area under the curve (AUC) of the percentage of correct vertices (PCV) and keypoints (PCK) follow~\cite{zimmermann2019freihand}. For object pose estimation, we report the percentage of average object 3D vertices error within $10\%$ of object diameter (ADD-0.1D). 

\subsection{Pose Estimation Performance}
\label{exp: pose-estimate}

\begin{table}[t]
    \tiny
    \centering
    \tablestyle{5pt}{1.05}
    \begin{tabular}{l|cccc|c}
      & \multicolumn{2}{c}{Hand Error($\downarrow$)} & \multicolumn{2}{c|}{F-score($\uparrow$)}
     & Object
     \\
     methods &  Joint &  Mesh &  F@5 &  F@15  & Estimation\\ 
     \shline
    Hasson~\textit{et al.}~\cite{hasson2019learning} & 11.1 & 11.0 & 46.0 & 93.0 & \cmark \\
    Hampali~\textit{et al.}~\cite{Hampali2019HOnnotateAM} & 10.7 & 10.6 & 50.6 & 94.2 & \xmark \\
    Ours & \textbf{9.9} & \textbf{9.5} & \textbf{52.6} & \textbf{95.5} & \cmark \\
    \end{tabular}
    \vspace{-0.1in}
    \caption{\small{Hand pose estimation performance compared with state-of-the-art methods on HO-3D~\cite{Hampali2019HOnnotateAM} dataset. The joint and mesh errors are in mm. The checkmark denotes whether the method also estimates the object pose. Our method has the lowest hand mesh error and also gives the object estimation simultaneously}}
    \label{tab: sota_compare}
    \vspace{-0.05in}
\end{table}

\begin{table}[t]
  \tablestyle{2.1pt}{1.05}
   \begin{tabular}{l|cccc|cccc}
    & \multicolumn{2}{c}{Hand Error($\downarrow$)} & \multicolumn{2}{c|}{F-score($\uparrow$)}
    & \multicolumn{4}{c}{Object ADD-0.1D($\uparrow$)}\\
    model & Joint & Mesh & F@5 & F@15 & cleanser & bottle & can & ave \\ \shline
    w/o CR & 10.2 & \textbf{9.7} & \textbf{53.7} & 94.9 & 75.3 & 59.1 & 52.1 & 62.2\\ 
    $\mathbf{h^+}$ & 10.3 & 9.8 & 52.7 & 94.9 & 85.4 & 67.8 & \textbf{54.4} & 69.2 \\
    $\mathbf{h^+o^+}$ & 10.7 & 10.3 & 51.6 & 94.3 & 86.9 & \textbf{69.6} & 53.2 & \textbf{69.9} \\
    $\mathbf{o^+}$ & \textbf{10.1} & \textbf{9.7} & 53.2 & \textbf{95.2} & \textbf{88.1} & 61.9 & 53.0 & 67.7 \\
    \end{tabular}
    \vspace{-0.1in}
    \caption{\small{Comparison of different queries in CR module on HO-3D dataset~\cite{Hampali2019HOnnotateAM}.
    $\mathbf{h^+}$, $\mathbf{h^+o^+}$ and $\mathbf{o^+}$ are to take hand features, both hand-object features, and object features as the query respectively. The key feature remains the same. 
    Ave means average. Performing contextual reasoning to enhance object representation (o+) could improve the object pose estimation, while using the hand features as query does not contribute to better hand pose estimation.}}
    \label{tab: cr_keq_query}
\vspace{-0.15in}
\end{table}

\textbf{Qualitative Results.}
The qualitative results on the HO-3D dataset~\cite{Hampali2019HOnnotateAM} are shown in Figure~\ref{fig:vis_ho3d}. We visualize the predicted hand-object in both 2D and 3D. It demonstrates our method can well handle the occlusion in the interaction and recover the accurate 3D hand mesh and object 6-Dof pose. Moreover, we visualize the cross-attention weights of different object query positions in the CR module in Figure~\ref{fig: vis_att}. We observe the CR module gives high responses to contact regions and tends to use the contact pattern for relational reasoning. 

\textbf{Comparison with State-of-the-Art.} We compare our hand pose estimation results with state-of-the-art-methods~\cite{hasson2019learning, Hampali2019HOnnotateAM} on the HO-3D dataset~\cite{Hampali2019HOnnotateAM} as shown in the Table~\ref{tab: sota_compare} and Figure~\ref{fig: curve_hand}. As can be seen from the figure, our method achieves the highest mesh-AUC at $80.9\%$, $1.9\%$, and $3.6\%$ higher than \cite{Hampali2019HOnnotateAM} and \cite{hasson2019learning} respectively. Our model also has the lowest hand mesh error of $9.5$mm. Besides the better hand pose performance, compared with\cite{hasson2019learning, Hampali2019HOnnotateAM}, our method could also give a much accurate object estimation simultaneously.

\subsection{Ablation Study}
\label{exp-ablation}
We perform the ablation study on HO-3D dataset~\cite{Hampali2019HOnnotateAM}. 
We first compare different designs of the CR module. Then, we investigate the effect of different filtering constraints and fractions of pseudo-labels contributed to hand and object pose estimation respectively.

\subsubsection{Ablation on CR Module}
\label{sec: ablation_cr}

We study the effect of different query design choices in the CR module under supervised learning. We compare three types of queries, while the key remains the same as the feature from the intersection region between hand-object. The first choice is to take the hand features as query and the CR module output is fused into the hand decoder, denoted as $\mathbf{h^+}$. The second choice is to take both the hand and object features as queries using two separate CR modules and the output is fused into both decoders, denoted as $\mathbf{h^+o^+}$. While the third one is what we proposed in Section~\ref{sec:cr} where we take object features as query, denoted as $\mathbf{o^+}$. As shown in Table~\ref{tab: cr_keq_query}, performing contextual reasoning to enhance object representation ($\mathbf{o^+}$) can improve the object pose estimation significantly. The average object ADD-0.1D has a $5.5\%$ improvement against the baseline without the CR module. However, using the hand features as a query ($\mathbf{h^+}$ and $\mathbf{h^+o^+}$) does not contribute to better hand pose estimation. It even degrades the performance against the baseline. This might be due to the CR module degenerates and the occlusion regions even distract the network attention.

\begin{table}[t]
\tiny
\centering
\tablestyle{2.1pt}{1.05}
\begin{tabular}{l|cccc|cccc}
& \multicolumn{2}{c}{Hand Error($\downarrow$)} & \multicolumn{2}{c|}{F-score($\uparrow$)}
& \multicolumn{4}{c}{Object ADD-0.1D($\uparrow$)}\\
model & Joint & Mesh & F@5 & F@15 & cleanser & bottle & can & ave \\ \shline
sup-w/o CR & 10.2 & \textbf{\textcolor{tblue}{9.7}} & \textbf{\textcolor{tblue}{53.7}} & 94.9 & 75.3 & 59.1 & 52.1 & 62.2\\ 
semi-w/o CR & 10.1 & 9.6 & \textbf{\textcolor{tred}{53.7}} & 95.2 & 88.8 & 68.9 & 49.8 & 69.2\\\hline
sup- w/ CR & \textbf{\textcolor{tblue}{10.1}} & \textbf{\textcolor{tblue}{9.7}} & 53.2 & \textbf{\textcolor{tblue}{95.2}} & \textbf{\textcolor{tblue}{88.1}} & \textbf{\textcolor{tblue}{61.9}} & \textbf{\textcolor{tblue}{53.0}} & \textbf{\textcolor{tblue}{67.7}}\\
semi- w/ CR & \textbf{\textcolor{tred}{9.9}} & \textbf{\textcolor{tred}{9.5}} & 52.6	 & \textbf{\textcolor{tred}{95.5}} & \textbf{\textcolor{tred}{92.2}} & \textbf{\textcolor{tred}{80.4}} & \textbf{\textcolor{tred}{55.7}} & \textbf{\textcolor{tred}{76.1}}\\
\end{tabular}
\vspace{-0.1in}
\caption{\small{Ablation analysis of semi-supervised learning on HO-3D dataset~\cite{Hampali2019HOnnotateAM}. \textbf{sup} and \textbf{semi} means under the supervised and semi-supervised phase. \textbf{w/ CR}  and \textbf{w/o CR} is the model with or without CR module. Best numbers of supervised learning results and semi-supervised learning results are shown in \textcolor{tblue}{blue} and \textcolor{tred}{red}. Ave means average. Semi-supervised learning with CR module gives the best hand-object estimation performance. }}
\label{tab: cr_ablation}
\end{table}

\begin{table}[t]
\tiny
\tablestyle{5pt}{1.05}
\begin{tabular}{l|c|cccc}
& Number & \multicolumn{2}{c}{Hand Error($\downarrow$)} & \multicolumn{2}{c}{F-score($\uparrow$)}\\
model & Pseudo-labels & Joint & Mesh & F@5 & F@15 \\ \shline
sup & 0 & 10.1 & 9.7 & \textbf{53.2} & 95.2 \\
w/o spatial  & 91,415  & 10.3 & 9.9 & 50.5 & 95.0 \\
w/o temporal & 91,415  & 10.3 & 9.8 & 51.6 & 95.1 \\
proposed & 91,415 & \textbf{9.9} & \textbf{9.5} & 52.6 & \textbf{95.5} \\
\end{tabular}
\vspace{-0.1in}
\caption{\small{Ablation analysis of using different filtering constraints in semi-supervised learning on the HO-3D dataset. Both constraints are critical for selecting high-quality pseudo-labels}}
\vspace{-0.1in}
\label{tab: constraint_ablation}
\end{table}

\subsubsection{Ablation on Semi-supervised Learning}

\textbf{Semi-supervised Learning Performance.} As shown in Table~\ref{tab: cr_ablation}, semi-supervised learning improves both hand and object pose estimation with or without using the CR module. 
Even though only hand pseudo-label are collected, the object pose estimation is benefited. We conjecture there are two reasons: First, better hand representation acts as a better context and could improve the object pose estimation via contextual reasoning; Second, the diversity of the hand pseudo-hand labels contribute to a much more robust shared backbone, which extracts base features for obtaining both hand and object representations via RoI Align.
With semi-supervised learning, the object ADD-0.1D of cleanser, bottle, and can are improved by $4.1\%$, $18.5\%$, $2.7\%$ respectively, as an $8.4\%$ improvement in average.

\begin{figure}[t]
\centering
  \begin{minipage}{0.23\textwidth}
  \includegraphics[width=4cm, height=3cm]{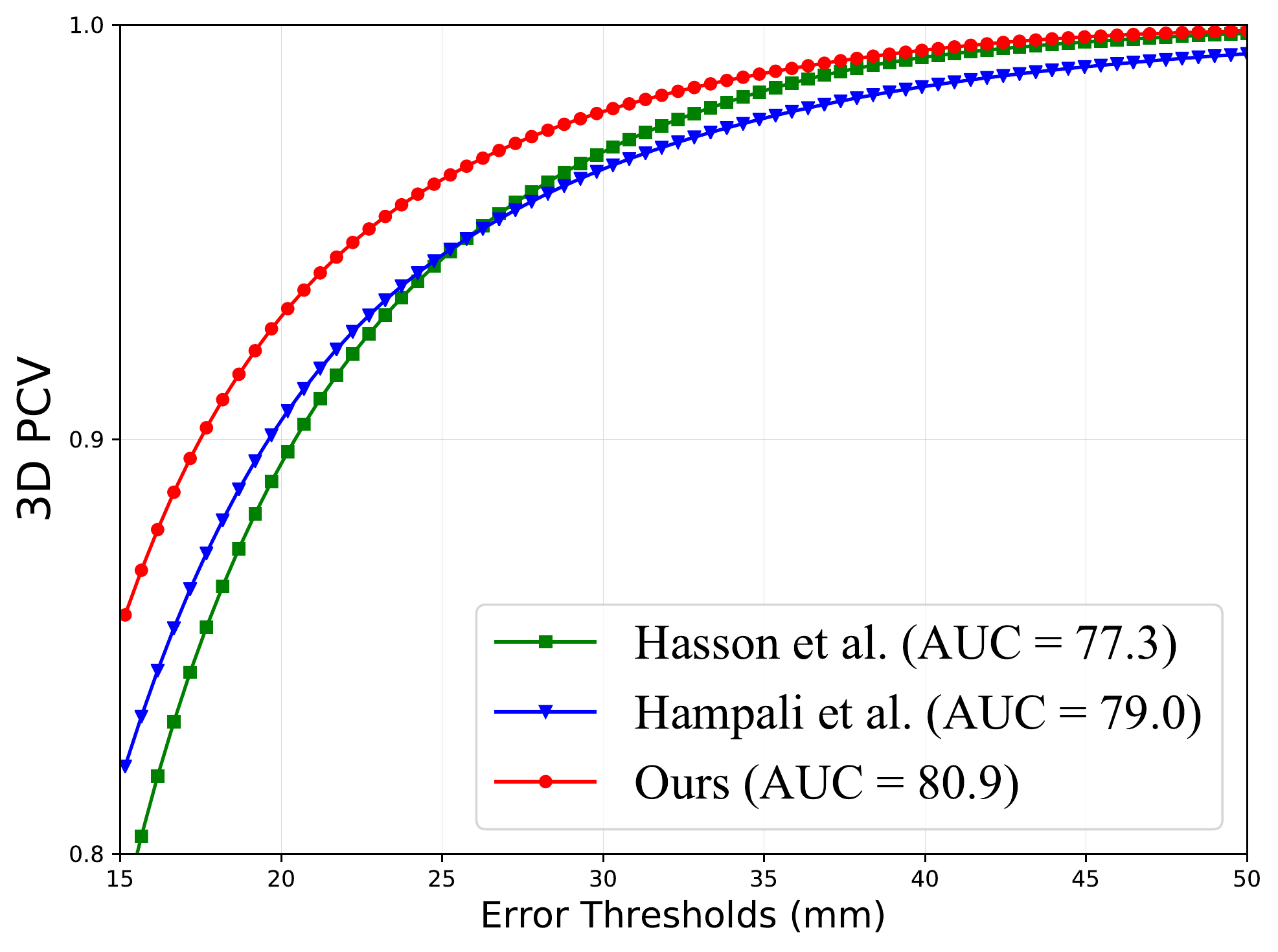}
  \vspace{-0.2in}
  \caption{\small{The hand mesh 3D PCV performance on the HO-3D dataset. Our method outperforms previous work~\cite{hasson2019learning, Hampali2019HOnnotateAM} by a large margin.}}
  \label{fig: curve_hand}
  \end{minipage}
  \hfill
  \begin{minipage}{0.23\textwidth}
  \includegraphics[width=4cm, height=3cm]{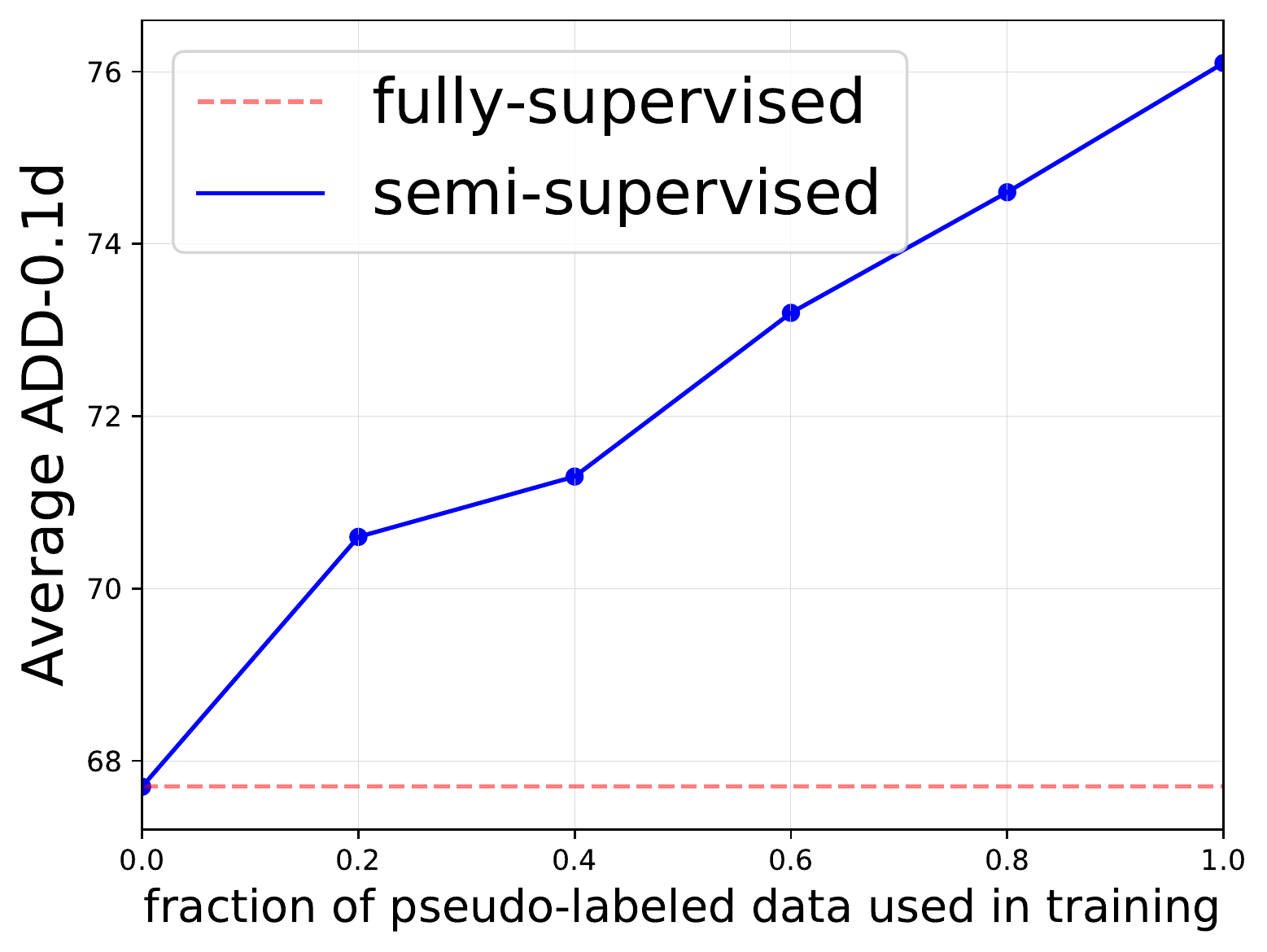}
  \caption{\small{Ablation study of different fractions of pseudo labels contributes to object pose performance gain on the HO-3D dataset.}}
  \label{fig: curve_object}
  \end{minipage}
\end{figure}

\begin{table}[t]
\tiny
\renewcommand{\arraystretch}{1.2}
\centering
\tablestyle{3pt}{1.05}
\begin{tabular}{l|cccc|cccc}
& \multicolumn{4}{c|}{FPHA~\cite{hasson2019learning}} & \multicolumn{4}{c}{Freihand~\cite{zimmermann2019freihand}}\\
& \multicolumn{2}{c}{AUC ($\uparrow$)} & \multicolumn{2}{c|}{F-Score ($\uparrow$)} & 
\multicolumn{2}{c}{AUC ($\uparrow$)} & \multicolumn{2}{c}{F-Score ($\uparrow$)} \\
\diagbox[width=7em]{model}{eval} & Joint & Mesh& F@5& F@15 & Joint & Mesh& F@5 & F@15 \\ \shline
supervised  & 74.1 & 74.6 & 43.6 & 91.7 &  69.1	& 68.2 & 34.8 & 85.3\\
semi-supervised & \textbf{75.4} & \textbf{76.0} & \textbf{45.7} & \textbf{93.1} & \textbf{70.6} & \textbf{69.6} & \textbf{36.3} & \textbf{86.6}\\
\end{tabular}
\vspace{-0.1in}
\caption{\small{Quantitative comparison of the cross-dataset generalization performance on FPHA~\cite{GarciaHernando2018FirstPersonHA} and Freihand~\cite{zimmermann2019freihand} dataset. The performance is measured as joint AUC, mesh AUC, and F-scores between models trained with supervised learning and semi-supervised learning. The model trained with semi-supervised has much better generalization performance.}}
\label{tab: cross_dataset_gen}
\vspace{-0.15in}
\end{table}

\textbf{Pseudo-labels Filtering Constraints.} 
We evaluate how the different filtering constraints contribute to the semi-supervised learning in Table~\ref{tab: constraint_ablation}. We compare against the method with only supervised learning and methods that remove the spatial or temporal filtering constraints. From the table, We can see each constraint plays an important role. Without either spatial or temporal constraints, it degrades the hand pose estimation accuracy. Therefore, both constraints are critical for selecting high-quality pseudo-labels and improving hand pose estimation performance.

\begin{figure*}[t]
\centering
\begin{tabular}{c@{\hskip 1.0in}c@{\hskip 2.4in}c@{\hskip 0.6in}c}
& FPHA~\cite{GarciaHernando2018FirstPersonHA} &  & Freihand~\cite{zimmermann2019freihand}
\end{tabular}
\includegraphics[width=17.5cm]{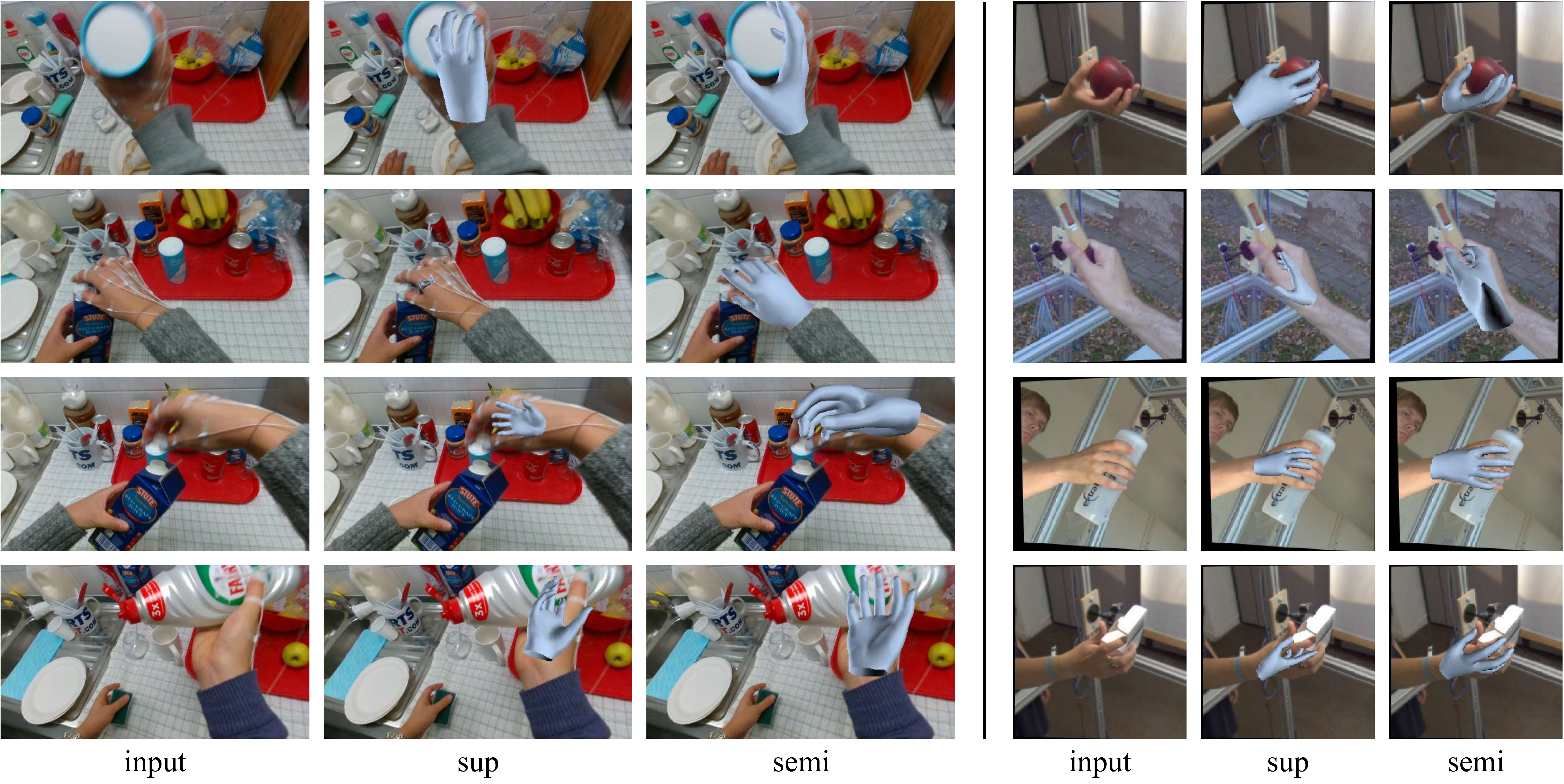}
\vspace{-0.3in}
\caption{\small{Qualitative comparison of the cross-dataset generalization performance on the FPHA~\cite{GarciaHernando2018FirstPersonHA} and Freihand~\cite{zimmermann2019freihand} dataset. The columns show the input hand-object interaction images, hand prediction results from models trained with supervised learning (sup) and semi-supervised learning (semi). All the testing models have not trained on the FPHA and Freihand dataset. The model trained with semi-supervised learning has a much more accurate hand estimation across different backgrounds, viewpoints, and subjects than the model trained only with supervised learning.}}
\label{fig: vis_gen}
\end{figure*}

\textbf{Amount of Pseudo-label.}
We analyze the effect of using different fractions of pseudo-labels in semi-supervised learning on the object ADD-0.1D performance on HO-3D dataset~\cite{Hampali2019HOnnotateAM}. We uniformly sample $20\%$, $40\%$, $60\%$, and $80\%$ fraction of the collected pseudo-labels for semi-supervised learning. As shown in Figure~\ref{fig: curve_object}, the more pseudo hand labels used in training, the better object performance the model could achieve. Even though the amount of annotated object pose keeps the same in training, the object performance could also be improved for two reasons: First, the better hand performance can help the object pose estimation via contextual reasoning explicitly; Second, the shared encoder is strengthened after semi-supervised learning and boosts object pose implicitly.

\subsection{Cross-domain Generalization Ability}
\label{exp-gen}
We evaluate the generalization performance of our model on FPHA~\cite{GarciaHernando2018FirstPersonHA} and Freihand~\cite{zimmermann2019freihand} dataset before and after semi-supervised learning. Follow the protocol of~\cite{zhang2019end, zimmermann2019freihand}, we report the joint AUC and mesh AUC after Procrustes alignment and  F-scores~\cite{knapitsch2017tanks}. To evaluate the hand mesh performance on the FPHA dataset, we fit the MANO model to the provided ground-truth hand joints following~\cite{hasson2020leveraging}. The quantitative and qualitative generalization results are shown in Table~\ref{tab: cross_dataset_gen} and Figure~\ref{fig: vis_gen}. As can be seen from the table and the figure, compared with the model trained only with supervised learning, the model trained with semi-supervised learning has much more accurate hand joints and mesh estimation across different backgrounds, viewpoints, and subjects. In semi-supervised learning, we utilize more training data from the Something-Something video dataset~\cite{Goyal2017TheS} that covers diverse hand poses interacting with objects and subjects in the wild. The model could thus benefit from those data sources in the semi-supervised training stage and yield much better results on generalization across different out-of-domain datasets. 

\vspace{-0.05in}
\section{Conclusion}
In this work, we propose a semi-supervised learning framework for estimating the 3D hand pose and 6-Dof object pose simultaneously, where the hand-object interaction regions are taken as the context for reasoning the object pose via the Transformer. 
After training the model on the annotated dataset, we deploy it on a large-scale video dataset to generate pseudo hand labels, and then perform spatial-temporal filtering to obtain high-quality ones. 
Finally, the model is retrained on the union set of real- and pseudo-labels under semi-supervised learning. Experimental results show that our method substantially improves the hand and object pose performance as well as has better cross-domain generalization.

{\footnotesize \textbf{Acknowledgements.}~This work was supported, in part, by grants from DARPA LwLL, NSF 1730158 CI-New: Cognitive Hardware and Software Ecosystem Community Infrastructure (CHASE-CI), NSF ACI-1541349 CC*DNI Pacific Research Platform, and gifts from Qualcomm and TuSimple.}

{\small
\bibliographystyle{ieee_fullname}
\bibliography{egbib}

\begin{thebibliography}{10}\itemsep=-1pt

\bibitem{arazo2019pseudo}
Eric Arazo, Diego Ortego, Paul Albert, Noel~E O'Connor, and Kevin McGuinness.
\newblock Pseudo-labeling and confirmation bias in deep semi-supervised
  learning.
\newblock {\em arXiv preprint arXiv:1908.02983}, 2019.

\bibitem{ba2016layer}
Jimmy~Lei Ba, Jamie~Ryan Kiros, and Geoffrey~E Hinton.
\newblock Layer normalization.
\newblock {\em arXiv preprint arXiv:1607.06450}, 2016.

\bibitem{baek2019pushing}
Seungryul Baek, Kwang~In Kim, and Tae-Kyun Kim.
\newblock Pushing the envelope for rgb-based dense 3d hand pose estimation via
  neural rendering.
\newblock In {\em CVPR}, pages 1067--1076, 2019.

\bibitem{battaglia2016interaction}
Peter Battaglia, Razvan Pascanu, Matthew Lai, Danilo~Jimenez Rezende, et~al.
\newblock Interaction networks for learning about objects, relations and
  physics.
\newblock In {\em Advances in Neural Information Processing Systems}, pages
  4502--4510, 2016.

\bibitem{boukhayma20193d}
Adnane Boukhayma, Rodrigo~de Bem, and Philip~HS Torr.
\newblock 3d hand shape and pose from images in the wild.
\newblock In {\em CVPR}, pages 10843--10852, 2019.

\bibitem{cai2018weakly}
Yujun Cai, Liuhao Ge, Jianfei Cai, and Junsong Yuan.
\newblock Weakly-supervised 3d hand pose estimation from monocular rgb images.
\newblock In {\em ECCV}, pages 666--682, 2018.

\bibitem{carion2020end}
Nicolas Carion, Francisco Massa, Gabriel Synnaeve, Nicolas Usunier, Alexander
  Kirillov, and Sergey Zagoruyko.
\newblock End-to-end object detection with transformers.
\newblock In {\em European Conference on Computer Vision}, pages 213--229,
  2020.

\bibitem{chen2013neil}
Xinlei Chen, Abhinav Shrivastava, and Abhinav Gupta.
\newblock Neil: Extracting visual knowledge from web data.
\newblock In {\em ICCV}, pages 1409--1416, 2013.

\bibitem{Chen2020JointH3}
Y. Chen, Z. Tu, Di Kang, Ruizhi Chen, Linchao Bao, Zhengyou Zhang, and J. Yuan.
\newblock Joint hand-object 3d reconstruction from a single image with
  cross-branch feature fusion.
\newblock {\em ArXiv}, abs/2006.15561, 2020.

\bibitem{chi2020relationnet++}
Cheng Chi, Fangyun Wei, and Han Hu.
\newblock Relationnet++: Bridging visual representations for object detection
  via transformer decoder.
\newblock {\em NeurIPS}, 2020.

\bibitem{choi2017robust}
Chiho Choi, Sang Ho~Yoon, Chin-Ning Chen, and Karthik Ramani.
\newblock Robust hand pose estimation during the interaction with an unknown
  object.
\newblock In {\em ICCV}, pages 3123--3132, 2017.

\bibitem{Doosti2020HOPENetAG}
Bardia Doosti, Shujon Naha, M. Mirbagheri, and David~J. Crandall.
\newblock Hope-net: A graph-based model for hand-object pose estimation.
\newblock {\em CVPR}, pages 6607--6616, 2020.

\bibitem{dosovitskiy2020image}
Alexey Dosovitskiy, Lucas Beyer, Alexander Kolesnikov, Dirk Weissenborn,
  Xiaohua Zhai, Thomas Unterthiner, Mostafa Dehghani, Matthias Minderer, Georg
  Heigold, Sylvain Gelly, et~al.
\newblock An image is worth 16x16 words: Transformers for image recognition at
  scale.
\newblock {\em arXiv preprint arXiv:2010.11929}, 2020.

\bibitem{garcia2020physics}
Guillermo Garcia-Hernando, Edward Johns, and Tae-Kyun Kim.
\newblock Physics-based dexterous manipulations with estimated hand poses and
  residual reinforcement learning.
\newblock {\em arXiv preprint arXiv:2008.03285}, 2020.

\bibitem{GarciaHernando2018FirstPersonHA}
Guillermo Garcia-Hernando, Shanxin Yuan, Seungryul Baek, and Tae-Kyun Kim.
\newblock First-person hand action benchmark with rgb-d videos and 3d hand pose
  annotations.
\newblock {\em CVPR}, pages 409--419, 2018.

\bibitem{ge20193d}
Liuhao Ge, Zhou Ren, Yuncheng Li, Zehao Xue, Yingying Wang, Jianfei Cai, and
  Junsong Yuan.
\newblock 3d hand shape and pose estimation from a single rgb image.
\newblock In {\em CVPR}, pages 10833--10842, 2019.

\bibitem{glauser2019interactive}
Oliver Glauser, Shihao Wu, Daniele Panozzo, Otmar Hilliges, and Olga
  Sorkine-Hornung.
\newblock Interactive hand pose estimation using a stretch-sensing soft glove.
\newblock {\em TOG}, 38(4):1--15, 2019.

\bibitem{Goyal2017TheS}
Raghav Goyal, Samira~Ebrahimi Kahou, Vincent Michalski, Joanna Materzynska,
  Susanne Westphal, Heuna Kim, Valentin Haenel, Ingo Fr{\"u}nd, Peter Yianilos,
  Moritz Mueller-Freitag, Florian Hoppe, Christian Thurau, Ingo Bax, and Roland
  Memisevic.
\newblock The “something something” video database for learning and
  evaluating visual common sense.
\newblock {\em ICCV}, pages 5843--5851, 2017.

\bibitem{Greff2017LSTMAS}
Klaus Greff, R. Srivastava, J. Koutn{\'i}k, Bastiaan Steunebrink, and J.
  Schmidhuber.
\newblock Lstm: A search space odyssey.
\newblock {\em IEEE Transactions on Neural Networks and Learning Systems},
  28:2222--2232, 2017.

\bibitem{Hampali2019HOnnotateAM}
Shreyas Hampali, Mahdi Rad, Markus Oberweger, and Vincent Lepetit.
\newblock Honnotate: A method for 3d annotation of hand and objects poses.
\newblock {\em CVPR}, 2019.

\bibitem{hampali2021handsformer}
Shreyas Hampali, Sayan~Deb Sarkar, Mahdi Rad, and Vincent Lepetit.
\newblock Handsformer: Keypoint transformer for monocular 3d pose estimation
  ofhands and object in interaction.
\newblock {\em arXiv preprint arXiv:2103.15320}, 2021.

\bibitem{handa2020dexpilot}
Ankur Handa, Karl Van~Wyk, Wei Yang, Jacky Liang, Yu-Wei Chao, Qian Wan, Stan
  Birchfield, Nathan Ratliff, and Dieter Fox.
\newblock Dexpilot: Vision-based teleoperation of dexterous robotic hand-arm
  system.
\newblock In {\em 2020 IEEE International Conference on Robotics and Automation
  (ICRA)}, pages 9164--9170. IEEE, 2020.

\bibitem{hasson2020leveraging}
Yana Hasson, Bugra Tekin, Federica Bogo, Ivan Laptev, Marc Pollefeys, and
  Cordelia Schmid.
\newblock Leveraging photometric consistency over time for sparsely supervised
  hand-object reconstruction.
\newblock {\em CVPR}, 2020.

\bibitem{hasson2019learning}
Yana Hasson, Gul Varol, Dimitrios Tzionas, Igor Kalevatykh, Michael~J Black,
  Ivan Laptev, and Cordelia Schmid.
\newblock Learning joint reconstruction of hands and manipulated objects.
\newblock In {\em CVPR}, pages 11807--11816, 2019.

\bibitem{he2017mask}
Kaiming He, Georgia Gkioxari, Piotr Doll{\'a}r, and Ross Girshick.
\newblock Mask r-cnn.
\newblock In {\em CVPR}, pages 2961--2969, 2017.

\bibitem{He2016DeepRL}
Kaiming He, Xiangyu Zhang, Shaoqing Ren, and Jian Sun.
\newblock Deep residual learning for image recognition.
\newblock {\em CVPR}, pages 770--778, 2016.

\bibitem{Hinton2015DistillingTK}
Geoffrey~E. Hinton, Oriol Vinyals, and J. Dean.
\newblock Distilling the knowledge in a neural network.
\newblock {\em ArXiv}, abs/1503.02531, 2015.

\bibitem{hu2018relation}
Han Hu, Jiayuan Gu, Zheng Zhang, Jifeng Dai, and Yichen Wei.
\newblock Relation networks for object detection.
\newblock In {\em Proceedings of the IEEE Conference on Computer Vision and
  Pattern Recognition}, pages 3588--3597, 2018.

\bibitem{Hu2019SegmentationDriven6O}
Yinlin Hu, Joachim Hugonot, Pascal Fua, and Mathieu Salzmann.
\newblock Segmentation-driven 6d object pose estimation.
\newblock {\em CVPR}, pages 3380--3389, 2019.

\bibitem{huang2020hand}
Lin Huang, Jianchao Tan, Ji Liu, and Junsong Yuan.
\newblock Hand-transformer: Non-autoregressive structured modeling for 3d hand
  pose estimation.
\newblock In {\em European Conference on Computer Vision}, pages 17--33, 2020.

\bibitem{hurst2013gesture}
Wolfgang H{\"u}rst and Casper Van~Wezel.
\newblock Gesture-based interaction via finger tracking for mobile augmented
  reality.
\newblock {\em Multimedia Tools and Applications}, 62(1):233--258, 2013.

\bibitem{iqbal2018hand}
Umar Iqbal, Pavlo Molchanov, Thomas Breuel Juergen~Gall, and Jan Kautz.
\newblock Hand pose estimation via latent 2.5 d heatmap regression.
\newblock In {\em ECCV}, pages 118--134, 2018.

\bibitem{iscen2019label}
Ahmet Iscen, Giorgos Tolias, Yannis Avrithis, and Ondrej Chum.
\newblock Label propagation for deep semi-supervised learning.
\newblock In {\em CVPR}, pages 5070--5079, 2019.

\bibitem{Kehl2017SSD6DMR}
Wadim Kehl, Fabian Manhardt, Federico Tombari, Slobodan Ilic, and Nassir Navab.
\newblock Ssd-6d: Making rgb-based 3d detection and 6d pose estimation great
  again.
\newblock {\em ICCV}, pages 1530--1538, 2017.

\bibitem{knapitsch2017tanks}
Arno Knapitsch, Jaesik Park, Qian-Yi Zhou, and Vladlen Koltun.
\newblock Tanks and temples: Benchmarking large-scale scene reconstruction.
\newblock {\em ACM Transactions on Graphics (ToG)}, 36(4):1--13, 2017.

\bibitem{dkulon2020cvpr}
Dominik Kulon, Riza~Alp Guler, Iasonas Kokkinos, Michael~M. Bronstein, and
  Stefanos Zafeiriou.
\newblock Weakly-supervised mesh-convolutional hand reconstruction in the wild.
\newblock In {\em CVPR}, June 2020.

\bibitem{lee2013pseudo}
Dong-Hyun Lee.
\newblock Pseudo-label: The simple and efficient semi-supervised learning
  method for deep neural networks.
\newblock In {\em Workshop on challenges in representation learning, ICML},
  volume~3, page~2, 2013.

\bibitem{lee2019set}
Juho Lee, Yoonho Lee, Jungtaek Kim, Adam Kosiorek, Seungjin Choi, and Yee~Whye
  Teh.
\newblock Set transformer: A framework for attention-based
  permutation-invariant neural networks.
\newblock In {\em Proceedings of the 36th International Conference on Machine
  Learning}, pages 3744--3753, 2019.

\bibitem{li2020category}
Xiaolong Li, He Wang, Li Yi, Leonidas~J Guibas, A~Lynn Abbott, and Shuran Song.
\newblock Category-level articulated object pose estimation.
\newblock In {\em Proceedings of the IEEE/CVF Conference on Computer Vision and
  Pattern Recognition}, pages 3706--3715, 2020.

\bibitem{lin2021end-to-end}
Kevin Lin, Lijuan Wang, and Zicheng Liu.
\newblock End-to-end human pose and mesh reconstruction with transformers.
\newblock In {\em CVPR}, 2021.

\bibitem{Lin2017FeaturePN}
Tsung-Yi Lin, Piotr Doll{\'a}r, Ross~B. Girshick, Kaiming He, Bharath
  Hariharan, and Serge~J. Belongie.
\newblock Feature pyramid networks for object detection.
\newblock {\em CVPR}, pages 936--944, 2017.

\bibitem{9028217}
Shaowei Liu, Guijin Wang, Pengwei Xie, and Cairong Zhang.
\newblock Light and fast hand pose estimation from spatial-decomposed latent
  heatmap.
\newblock {\em IEEE Access}, 8:53072--53081, 2020.

\bibitem{lopez2015unifying}
David Lopez-Paz, L{\'e}on Bottou, Bernhard Sch{\"o}lkopf, and Vladimir Vapnik.
\newblock Unifying distillation and privileged information.
\newblock {\em arXiv preprint arXiv:1511.03643}, 2015.

\bibitem{mao2021tfpose}
Weian Mao, Yongtao Ge, Chunhua Shen, Zhi Tian, Xinlong Wang, and Zhibin Wang.
\newblock Tfpose: Direct human pose estimation with transformers.
\newblock {\em arXiv preprint arXiv:2103.15320}, 2021.

\bibitem{moon2018v2v}
Gyeongsik Moon, Ju~Yong Chang, and Kyoung~Mu Lee.
\newblock V2v-posenet: Voxel-to-voxel prediction network for accurate 3d hand
  and human pose estimation from a single depth map.
\newblock In {\em Proceedings of the IEEE conference on computer vision and
  pattern Recognition}, pages 5079--5088, 2018.

\bibitem{GANeratedHands_CVPR2018}
Franziska Mueller, Florian Bernard, Oleksandr Sotnychenko, Dushyant Mehta,
  Srinath Sridhar, Dan Casas, and Christian Theobalt.
\newblock Ganerated hands for real-time 3d hand tracking from monocular rgb.
\newblock In {\em CVPR}, page~11, 2018.

\bibitem{mueller2017real}
Franziska Mueller, Dushyant Mehta, Oleksandr Sotnychenko, Srinath Sridhar, Dan
  Casas, and Christian Theobalt.
\newblock Real-time hand tracking under occlusion from an egocentric rgb-d
  sensor.
\newblock In {\em ICCVW}, pages 1284--1293, 2017.

\bibitem{newell2016stacked}
Alejandro Newell, Kaiyu Yang, and Jia Deng.
\newblock Stacked hourglass networks for human pose estimation.
\newblock In {\em ECCV}, pages 483--499. Springer, 2016.

\bibitem{oberweger2019generalized}
Markus Oberweger, Paul Wohlhart, and Vincent Lepetit.
\newblock Generalized feedback loop for joint hand-object pose estimation.
\newblock {\em IEEE TPAMI}, 2019.

\bibitem{oikonomidis2011full}
Iason Oikonomidis, Nikolaos Kyriazis, and Antonis~A Argyros.
\newblock Full dof tracking of a hand interacting with an object by modeling
  occlusions and physical constraints.
\newblock In {\em ICCV}, pages 2088--2095. IEEE, 2011.

\bibitem{parmar2018image}
Niki Parmar, Ashish Vaswani, Jakob Uszkoreit, Lukasz Kaiser, Noam Shazeer,
  Alexander Ku, and Dustin Tran.
\newblock Image transformer.
\newblock In {\em International Conference on Machine Learning}, pages
  4055--4064, 2018.

\bibitem{pavlakos2018learning}
Georgios Pavlakos, Luyang Zhu, Xiaowei Zhou, and Kostas Daniilidis.
\newblock Learning to estimate 3d human pose and shape from a single color
  image.
\newblock In {\em Proceedings of the IEEE Conference on Computer Vision and
  Pattern Recognition}, pages 459--468, 2018.

\bibitem{Peng2019PVNetPV}
Sida Peng, Yuan Liu, Qi-Xing Huang, Hujun Bao, and Xiaowei Zhou.
\newblock Pvnet: Pixel-wise voting network for 6dof pose estimation.
\newblock {\em CVPR}, pages 4556--4565, 2019.

\bibitem{piumsomboon2013user}
Thammathip Piumsomboon, Adrian Clark, Mark Billinghurst, and Andy Cockburn.
\newblock User-defined gestures for augmented reality.
\newblock In {\em IFIP Conference on Human-Computer Interaction}, pages
  282--299, 2013.

\bibitem{Rad2017BB8AS}
Mahdi Rad and Vincent Lepetit.
\newblock Bb8: A scalable, accurate, robust to partial occlusion method for
  predicting the 3d poses of challenging objects without using depth.
\newblock {\em ICCV}, pages 3848--3856, 2017.

\bibitem{radosavovic2018data}
Ilija Radosavovic, Piotr Doll{\'a}r, Ross Girshick, Georgia Gkioxari, and
  Kaiming He.
\newblock Data distillation: Towards omni-supervised learning.
\newblock In {\em CVPR}, pages 4119--4128, 2018.

\bibitem{ramachandran2019stand}
Prajit Ramachandran, Niki Parmar, Ashish Vaswani, Irwan Bello, Anselm Levskaya,
  and Jonathon Shlens.
\newblock Stand-alone self-attention in vision models.
\newblock In {\em Advances in Neural Information Processing Systems}, 2019.

\bibitem{Redmon2016YouOL}
Joseph Redmon, Santosh~Kumar Divvala, Ross~B. Girshick, and Ali Farhadi.
\newblock You only look once: Unified, real-time object detection.
\newblock {\em CVPR}, pages 779--788, 2016.

\bibitem{riloff2003learning}
Ellen Riloff and Janyce Wiebe.
\newblock Learning extraction patterns for subjective expressions.
\newblock In {\em EMNLP}, pages 105--112, 2003.

\bibitem{rogez20143d}
Gr{\'e}gory Rogez, Maryam Khademi, JS Supan{\v{c}}i{\v{c}}~III, Jose
  Maria~Martinez Montiel, and Deva Ramanan.
\newblock 3d hand pose detection in egocentric rgb-d images.
\newblock In {\em ECCV}, pages 356--371, 2014.

\bibitem{romero2017embodied}
Javier Romero, Dimitrios Tzionas, and Michael~J Black.
\newblock Embodied hands: Modeling and capturing hands and bodies together.
\newblock {\em ToG}, 36(6):245, 2017.

\bibitem{rosenberg2005semi}
Chuck Rosenberg, Martial Hebert, and Henry Schneiderman.
\newblock Semi-supervised self-training of object detection models.
\newblock {\em WACV/MOTION}, 2, 2005.

\bibitem{sadeghian2017tracking}
Amir Sadeghian, Alexandre Alahi, and Silvio Savarese.
\newblock Tracking the untrackable: Learning to track multiple cues with
  long-term dependencies.
\newblock In {\em Proceedings of the IEEE International Conference on Computer
  Vision}, pages 300--311, 2017.

\bibitem{santoro2017simple}
Adam Santoro, David Raposo, David~G Barrett, Mateusz Malinowski, Razvan
  Pascanu, Peter Battaglia, and Timothy Lillicrap.
\newblock A simple neural network module for relational reasoning.
\newblock In {\em Advances in Neural Information Processing Systems},
  volume~30, 2017.

\bibitem{schwarz2015rgb}
Max Schwarz, Hannes Schulz, and Sven Behnke.
\newblock Rgb-d object recognition and pose estimation based on pre-trained
  convolutional neural network features.
\newblock In {\em 2015 IEEE international conference on robotics and automation
  (ICRA)}, pages 1329--1335. IEEE, 2015.

\bibitem{scudder1965probability}
H Scudder.
\newblock Probability of error of some adaptive pattern-recognition machines.
\newblock {\em IEEE Transactions on Information Theory}, 11(3):363--371, 1965.

\bibitem{shi2018transductive}
Weiwei Shi, Yihong Gong, Chris Ding, Zhiheng MaXiaoyu~Tao, and Nanning Zheng.
\newblock Transductive semi-supervised deep learning using min-max features.
\newblock In {\em ECCV}, pages 299--315, 2018.

\bibitem{spurr2018cvpr}
Adrian Spurr, Jie Song, Seonwook Park, and Otmar Hilliges.
\newblock Cross-modal deep variational hand pose estimation.
\newblock In {\em CVPR}, 2018.

\bibitem{sridhar2015investigating}
Srinath Sridhar, Anna~Maria Feit, Christian Theobalt, and Antti Oulasvirta.
\newblock Investigating the dexterity of multi-finger input for mid-air text
  entry.
\newblock In {\em CHI}, pages 3643--3652, 2015.

\bibitem{stoffl2021end}
Lucas Stoffl, Maxime Vidal, and Alexander Mathis.
\newblock End-to-end trainable multi-instance pose estimation with
  transformers.
\newblock {\em arXiv preprint arXiv:2103.12115}, 2021.

\bibitem{sun2019videobert}
Chen Sun, Austin Myers, Carl Vondrick, Kevin Murphy, and Cordelia Schmid.
\newblock Videobert: A joint model for video and language representation
  learning.
\newblock In {\em Proceedings of the IEEE/CVF International Conference on
  Computer Vision}, pages 7464--7473, 2019.

\bibitem{tekin2019h+}
Bugra Tekin, Federica Bogo, and Marc Pollefeys.
\newblock H+o: Unified egocentric recognition of 3d hand-object poses and
  interactions.
\newblock In {\em CVPR}, pages 4511--4520, 2019.

\bibitem{Tekin2018RealTimeSS}
Bugra Tekin, Sudipta~N. Sinha, and Pascal Fua.
\newblock Real-time seamless single shot 6d object pose prediction.
\newblock {\em CVPR}, pages 292--301, 2018.

\bibitem{vaswani2017attention}
Ashish Vaswani, Noam Shazeer, Niki Parmar, Jakob Uszkoreit, Llion Jones,
  Aidan~N Gomez, Lukasz Kaiser, and Illia Polosukhin.
\newblock Attention is all you need.
\newblock {\em NeurIPS}, page 5998–6008, 2017.

\bibitem{Wang2018NonlocalNN}
Xiaolong Wang, Ross~B. Girshick, Abhinav Gupta, and Kaiming He.
\newblock Non-local neural networks.
\newblock {\em CVPR}, pages 7794--7803, 2018.

\bibitem{wang2018videos}
Xiaolong Wang and Abhinav Gupta.
\newblock Videos as space-time region graphs.
\newblock In {\em Proceedings of the European conference on computer vision
  (ECCV)}, pages 399--417, 2018.

\bibitem{watters2017visual}
Nicholas Watters, Daniel Zoran, Theophane Weber, Peter Battaglia, Razvan
  Pascanu, and Andrea Tacchetti.
\newblock Visual interaction networks: Learning a physics simulator from video.
\newblock {\em Advances in neural information processing systems},
  30:4539--4547, 2017.

\bibitem{wu2019long}
Chao-Yuan Wu, Christoph Feichtenhofer, Haoqi Fan, Kaiming He, Philipp
  Krahenbuhl, and Ross Girshick.
\newblock Long-term feature banks for detailed video understanding.
\newblock In {\em Proceedings of the IEEE/CVF Conference on Computer Vision and
  Pattern Recognition}, pages 284--293, 2019.

\bibitem{Xiang2020RevisitingTC}
Sitao Xiang and Hao Li.
\newblock Revisiting the continuity of rotation representations in neural
  networks.
\newblock {\em ArXiv}, abs/2006.06234, 2020.

\bibitem{xiang2017posecnn}
Yu Xiang, Tanner Schmidt, Venkatraman Narayanan, and Dieter Fox.
\newblock Posecnn: A convolutional neural network for 6d object pose estimation
  in cluttered scenes.
\newblock {\em arXiv preprint arXiv:1711.00199}, 2017.

\bibitem{Xiang2018PoseCNNAC}
Yu Xiang, Tanner Schmidt, Venkatraman Narayanan, and Dieter Fox.
\newblock Posecnn: A convolutional neural network for 6d object pose estimation
  in cluttered scenes.
\newblock {\em ArXiv}, abs/1711.00199, 2018.

\bibitem{xie2019self}
Qizhe Xie, Eduard Hovy, Minh-Thang Luong, and Quoc~V Le.
\newblock Self-training with noisy student improves imagenet classification.
\newblock {\em arXiv preprint arXiv:1911.04252}, 2019.

\bibitem{xu2017lie}
Chi Xu, Lakshmi~Narasimhan Govindarajan, Yu Zhang, and Li Cheng.
\newblock Lie-x: Depth image based articulated object pose estimation,
  tracking, and action recognition on lie groups.
\newblock {\em International Journal of Computer Vision}, 123:454--478, 2017.

\bibitem{yang2020transpose}
Sen Yang, Zhibin Quan, Mu Nie, and Wankou Yang.
\newblock Transpose: Towards explainable human pose estimation by transformer.
\newblock {\em arXiv preprint arXiv:2012.14214}, 2020.

\bibitem{yao2012real}
Yuan Yao and Yun Fu.
\newblock Real-time hand pose estimation from rgb-d sensor.
\newblock In {\em ICME}, pages 705--710, 2012.

\bibitem{yarowsky1995unsupervised}
David Yarowsky.
\newblock Unsupervised word sense disambiguation rivaling supervised methods.
\newblock In {\em ACL}, pages 189--196, 1995.

\bibitem{yuan2018depth}
Shanxin Yuan, Guillermo Garcia-Hernando, Bj{\"o}rn Stenger, Gyeongsik Moon,
  Ju~Yong Chang, Kyoung~Mu Lee, Pavlo Molchanov, Jan Kautz, Sina Honari, Liuhao
  Ge, et~al.
\newblock Depth-based 3d hand pose estimation: From current achievements to
  future goals.
\newblock In {\em Proceedings of the IEEE Conference on Computer Vision and
  Pattern Recognition}, pages 2636--2645, 2018.

\bibitem{zhang2019end}
Xiong Zhang, Qiang Li, Hong Mo, Wenbo Zhang, and Wen Zheng.
\newblock End-to-end hand mesh recovery from a monocular rgb image.
\newblock In {\em CVPR}, pages 2354--2364, 2019.

\bibitem{zhou2020monocular}
Yuxiao Zhou, Marc Habermann, Weipeng Xu, Ikhsanul Habibie, Christian Theobalt,
  and Feng Xu.
\newblock Monocular real-time hand shape and motion capture using multi-modal
  data.
\newblock {\em CVPR}, 2020.

\bibitem{zimmermann2017learning}
Christian Zimmermann and Thomas Brox.
\newblock Learning to estimate 3d hand pose from single rgb images.
\newblock In {\em CVPR}, pages 4903--4911, 2017.

\bibitem{zimmermann2019freihand}
Christian Zimmermann, Duygu Ceylan, Jimei Yang, Bryan Russell, Max Argus, and
  Thomas Brox.
\newblock Freihand: A dataset for markerless capture of hand pose and shape
  from single rgb images.
\newblock In {\em Proceedings of the IEEE/CVF International Conference on
  Computer Vision}, pages 813--822, 2019.

\end{thebibliography}
}

\clearpage
\appendix
\pdfoutput=1

\setcounter{section}{0}

\begin{center}
\textbf{\Large Appendix}
\end{center}

\section*{Appendix A: Network architecture}
The network architecture is illustrated in Table~\ref{tab: h-o pose network}. 
We utilize ROIAlign~\cite{he2017mask} to crop the hand and the object features $\mathcal{F}_h$ and $\mathcal{F}_o$ with channel dimensions $256$ and spatial resolution $32$ from the most fine-grained level P2 features of the FPN backbone. 
The CR module enhances the query object features and maintains the feature size.

\textbf{Hand Decoder}
The 2D heatmaps $\mathcal{H}$ in the joints localization network takes the form of $\mathcal{H} \in \mathbb{R}^{32\times 32 \times N_h}$ where each channel corresponds to one joint. $N_h = 21$ is the number of joints. Then the 2D joint positions $\mathcal{J}^{2D} \in \mathcal{R}^{N_h \times2}$ can be calculated by the weighted sum of heatmap values and corresponding 2D pixel coordinates as $\mathcal{J}^{2D}_i = \sum p\cdot H_i(p)$ for each joint $i$, where $p$ represents the pixel coordinates in the heatmap. The mesh regression network predicts the MANO pose parameters $\theta \in \mathbb{R}^{48}$ and shape parameters $\beta \in \mathbb{R}^{10}$ by first using the residual blocks combined with max-pooling layers to downsample the hand features, and three fully-connected layers afterward to regress the target MANO parameters $\theta$ and $\beta$.

\textbf{Object Decoder}
The object decoder has two outputs, the control points 2D coordinates in $\mathbb{R}^{32 \times 32 \times N_o \times 2}$ and the corresponding 1D confidence values in $\mathbb{R}^{32 \times 32 \times N_o \times 1}$ for all grids, where $N_o=21$ is the number of control points. To predict the i-th control point pixel coordinates from grid $g$, the network estimates a offset $v_{g,i}$ between the grid's pixel location $p_g$ and the target control point position $t_i$. Then the residual error $\delta_{g, i}$ between the prediction in grid $g$ to the target control point $i$ is $\delta_{g, i}=p_g + v_{g,i} - t_i$. The confidence score $c_{g,i}$ is obtained by apply the sigmoid function at the top of the second stream output.

\begin{table}
\scriptsize
\tablestyle{7pt}{1.2}
\begin{tabular}{ccc}\shline
 Stage & Configuration & Output\\\shline
 0 & Input image & $512\times 512 \times 3$\\\shline
  & {\textbf{Feature Extraction}}  & \\ \hline
 \multirow{1}*{1} & \multirow{1}*{Res-50-FPN ($P_2$)~
 \cite{Lin2017FeaturePN}} & $128 \times 128 \times 256$\\ \hline
  \multirow{1}*{1} & \multirow{1}*{Hand-RoiAlign~\cite{he2017mask}} & $32\times 32\times 256$\\ \hline
  \multirow{1}*{1} & \multirow{1}*{Object-RoiAlign~\cite{he2017mask}} & $32 \times 32 \times 256$\\ \hline
& {\textbf{Contextual Reasoning}}  & \\ \hline
 \multirow{1}*{2} & enhanced object feature & $32\times32\times 256$ \\ \shline
  & {\textbf{Hand Decoder}}  & \\ \hline
 \multirow{1}*{3} & \multirow{1}*{Hourglass Module~\cite{newell2016stacked}} & $32\times 32\times 21$\\ \hline
 \multirow{1}*{3} & \multirow{1}*{4 Residaul Block~\cite{He2016DeepRL}} & $2\times 2\times 512$\\ \hline
 \multirow{1}*{3} & \multirow{1}*{Flatten} & $2048$\\ \hline
 \multirow{1}*{3} & \multirow{1}*{3 FC Layers} & $58$\\ \hline 
& {\textbf{Object Decoder}}  & \\ \hline
\multirow{1}*{4} & \multirow{1}*{4 Shared 2D Convolution} & $32\times32\times256$\\ \hline
\multirow{1}*{4} & \multirow{1}*{2D Convolution of localization} & $32\times32\times21\times2$\\ \hline
\multirow{1}*{4} & \multirow{1}*{2D Convolution of confidence} & $32\times32\times21\times1$\\ \hline
\end{tabular}
\vspace{-0.05in}
\caption{Network architecture and configurations of the proposed model. FC layers denote the fully-connected layers.}
\label{tab: h-o pose network}
\end{table}

\begin{table}
\scriptsize
\tablestyle{7pt}{1.2}
\begin{tabular}{l|c|c}
 & \multicolumn{1}{c|}{Hand}
 & \multicolumn{1}{c}{Object}
 \\
 models & \multicolumn{1}{c|}{mean distance ($\downarrow$)}
& \multicolumn{1}{c}{mean distance ($\downarrow$)}
 \\
\shline
Tekin \textit{et. al}~\cite{tekin2019h+}  & \textbf{15.8} & 24.9  \\\hline
Hasson \textit{et. al}~\cite{hasson2020leveraging}  & 18.0  & 22.3  \\\hline
\textbf{Ours-sup}  & 16.0  & \textbf{19.7}  \\ 
\end{tabular}
\vspace{-0.05in}
\caption{Hand pose and object pose performance between state-of-the-art methods~\cite{tekin2019h+, hasson2020leveraging} and the proposed method on FPHA~\cite{GarciaHernando2018FirstPersonHA} dataset. \textbf{sup} means our model is trained under the supervised learning phase. The error is in mm.}
\label{tabel: compare_fpha}
\end{table}

\begin{table}
\tiny
\tablestyle{5pt}{1.05}
\begin{tabular}{l|cccc}
& \multicolumn{2}{c}{Hand AUC($\uparrow$)} & \multicolumn{2}{c}{F-score($\uparrow$)}\\
methods & Joint & Mesh & F@5 & F@15 \\ \shline
Boukhayma~\textit{et al.}~\cite{boukhayma20193d} & 73.1 & 74.1 & 43.2 & 90.9\\
Radosavovic~\textit{et al.}~\cite{radosavovic2018data} & 74.9 & 75.4 & 44.8 & 92.5 \\
Ours & \textbf{75.4} & \textbf{76.0} & \textbf{45.7} &  \textbf{93.1} \\
\end{tabular}
\caption{cross-domain generalization comparison against other semi-supervised approaches on the FPHA dataset.}
\label{tab: semi_compare}
\vspace{-0.1in}
\end{table}

\section*{Appendix B: Performance on FPHA Dataset with Fully Supervision}
We report our method's hand and object pose estimation performance on FPHA~\cite{GarciaHernando2018FirstPersonHA}
dataset. Considering the significant appearance change between the FPHA dataset and other datasets because of the introduction of visible markers used for annotation, we do not use the FPHA dataset for semi-supervised learning. We compare our model's performance trained under supervised learning against other state-of-the-art approaches~\cite{tekin2019h+, hasson2020leveraging}, the results are shown in Table~\ref{tabel: compare_fpha}. As can be seen, our method has the lowest object estimation error and outperforms other approaches by a large margin, as well as a comparable hand pose estimation performance against~\cite{tekin2019h+} and much better than~\cite{hasson2020leveraging}. Our model with only supervised learning could achieve the best overall performance on the FPHA dataset, which demonstrates the superiority of the joint learning framework and the effectiveness of the contextual reasoning module.

\section*{Appendix C: Cross-domain Semi-supervised Learning Results on FPHA Dataset}

We compare our method with other semi-supervised learning methods~\cite{boukhayma20193d, radosavovic2018data} on the FPHA dataset as shown in Table.~\ref{tab: semi_compare}. Specifically, we implemented~\cite{radosavovic2018data} by removing the spatial-temporal constraints in our method for generating pseudo labels. Note that all the methods are not trained with the FPHA dataset. Our method performs significantly better than previous approaches in this cross-domain setting. Our method earns such benefits from using real-world hand-object videos. Training on these videos covers most domains in different test datasets. Thus we can improve the generalization across different domains and datasets. 

\end{document}